\documentclass[11pt]{article}

\usepackage[final]{acl}
\usepackage{tabularx}
\usepackage{colortbl} 

\newcolumntype{C}{>{\centering\arraybackslash}X}
\usepackage{times}
\usepackage{latexsym}
\usepackage[T1]{fontenc}

\usepackage{microtype}

\usepackage{inconsolata}
\usepackage{makecell}

\usepackage{enumitem}
\usepackage{amsmath,amsfonts}
\usepackage[utf8]{inputenc}
\usepackage{booktabs}
\usepackage{multirow}
\usepackage{graphicx}
\usepackage{pifont}
\usepackage{tabularx}
\usepackage[table]{xcolor} 
\usepackage[raster,skins,most]{tcolorbox}
\usepackage{fvextra}
\usepackage{algorithm}
\usepackage{algpseudocode}
\usepackage{fontawesome5}
\newcommand{\myemoji}[1]{\includegraphics[height=1em]{#1}}

\tcbset{
  geoprompt/.style={
    colback=gray!10,      
    colframe=darkgray,    
    coltitle=white,       
    arc=2mm,              
    fonttitle=\bfseries,  
  },
  geotool/.style={
    colback=gray!10,
    colframe=darkgray,
    coltitle=white,
    arc=2mm,
    fonttitle=\bfseries,
    boxrule=0.5pt,
    left=3pt,
    right=3pt,
    top=3pt,
    bottom=3pt,
  }
}

\definecolor{boxborder}{HTML}{2B5A84}
\definecolor{highlighttext}{HTML}{3A74A6}
\definecolor{errortext}{HTML}{C0392B}
\definecolor{graytext}{HTML}{7F8C8D}
\definecolor{oktext}{HTML}{2E7D32}
\definecolor{warntxt}{HTML}{B36B00}

\tcbset{
    headerbox/.style={
        colback=gray!7,
        colframe=boxborder,
        boxrule=0.8pt,
        arc=2pt,
        left=3pt, right=3pt, top=1.2pt, bottom=1.2pt,
        boxsep=0pt,
        height=0.42cm,
        valign=center,
        before skip=0pt,
        after skip=0.7mm,
        enhanced,
        drop fuzzy shadow={black!14!white}
    },
    modelbox/.style={
        colback=white,
        colframe=boxborder,
        boxrule=0.8pt,
        arc=2pt,
        left=4pt, right=4pt, top=3pt, bottom=3pt,
        boxsep=0pt,
        before skip=0pt,
        after skip=0pt,
        enhanced,
        drop fuzzy shadow={black!18!white}
    },
    modelboxtop/.style={
        modelbox,
        height=7.35cm,
        valign=top
    },
    imagecaptionbox/.style={
        colback=gray!7,
        colframe=boxborder,
        boxrule=0.8pt,
        arc=2pt,
        left=3pt, right=3pt, top=2pt, bottom=2pt,
        boxsep=0pt,
        before skip=0pt,
        after skip=0pt,
        enhanced,
        drop fuzzy shadow={black!14!white}
    },
    imageframebox/.style={
        colback=white,
        colframe=boxborder,
        boxrule=0.8pt,
        arc=2pt,
        left=0pt, right=0pt, top=0pt, bottom=0pt,
        boxsep=0pt,
        before skip=0pt,
        after skip=0pt,
        enhanced,
        drop fuzzy shadow={black!16!white}
    },
    modelboxbottom/.style={
        modelbox,
        height=3.55cm,
        valign=top
    },
    modelboxgpt/.style={
        modelbox,
        height=5.45cm,
        valign=top
    }
}

\newcommand{\modelheader}[3]{%
    \par\noindent
    \begin{tcolorbox}[headerbox]
        \makebox[\linewidth][s]{%
            \raisebox{-0.18em}{\includegraphics[height=0.95em]{#1}}\hspace{0.22em}%
            {\strut\sffamily\bfseries\scriptsize\textcolor{boxborder}{#2}}%
            \hfill%
            {\strut\sffamily\bfseries\scriptsize\textcolor{highlighttext}{#3}}%
        }%
    \end{tcolorbox}%
}
\newcommand{\imageheader}[1]{%
    \par\noindent
    \begin{tcolorbox}[headerbox]
        \centering{\strut\sffamily\bfseries\scriptsize\textcolor{highlighttext}{#1}}%
    \end{tcolorbox}%
}
\newcommand{\traceitem}[2]{\noindent{\sffamily\bfseries\scriptsize #1}\hspace{0.18em}{\scriptsize #2}\par\vspace{0.35mm}}

\title{Perceive to Hypothesize, Verify to Ground: An Agentic Reasoning Framework for Open-World Geo-Localization}

\author{
  Yutian Jiang$^{1*}$,
  Ruijie Li$^{1*}$,
   Sisuo Lyu$^{2*}$,
  Xixuan Hao$^{1}$, \\
  \textbf{Qingxiang Liu}$^{1}$,
  \textbf{Yongzi Yu}$^{1}$,
  \textbf{Yuxuan Liang}$^{1\dagger}$ \\
  $^{1}$The Hong Kong University of Science and Technology (Guangzhou) \\
  $^{2}$The Hong Kong University of Science and Technology \\
  \texttt{\{yjiang194, rli541, xhao390, yyu110\}@connect.hkust-gz.edu.cn} \\
  \texttt{\{sisuolyu, yuxliang\}@outlook.com}, \texttt{qingxiangliu737@gmail.com}
}

\begin{document}
\maketitle

\begingroup
\renewcommand{\thefootnote}{\fnsymbol{footnote}}
\footnotetext[1]{Equal contribution.}
\footnotetext[2]{Corresponding author.}
\endgroup

\begin{abstract}

Open-world geo-localization requires models to reason over ambiguous visual cues through multi-step reasoning and external knowledge grounding. While recent large vision-language models exhibit strong multimodal reasoning capabilities, existing approaches still suffer from perceptual hallucination and context drift due to the lack of explicit evidence-grounded verification. In this work, we reformulate geo-localization as a human-like \textbf{perceive-then-verify} reasoning problem and propose \textbf{GeoPAVE} (Geo-localization Perception-and-Verification-Engine), a bi-level agentic framework that contains perception-based hypothesis generation via single-pass rollouts and verification-based evidence grounding for decision actions: \textit{support}, \textit{refute}, and \textit{refine}. To support rigorous evaluation, we further introduce \textbf{PAVED}, a novel dataset derived from real-world user check-in data, equipped with comprehensive reasoning trajectories featuring multi-hop queries, multi-round tool invocations, and structured perception-verification traces. The dataset and code are available at https://github.com/Arandinglv/GeoPAVE.
\end{abstract}

\section{Introduction}
Open-world geo-localization aims to infer the geographic coordinates of a street-view image around the world, without relying on a retrieval gallery. Therefore, this task requires models to integrate fine-grained visual evidence~\cite{urbanvlp,satcle,zhong2024urbancross} with broad geographic, cultural, and semantic knowledge~\cite{yang2026skillgeo,jia2025geoarena,hao2025geospatial,zou2025deep}. This moves past simple single-shot perception, requiring iterative reasoning, multi-hop retrieval, and fact verification with open-world exploration over external knowledge sources~\cite{cai2026mata,pattern-matching}.  Geo-localization supports a wide range of real-world applications, such as autonomous driving, criminal investigation~\cite{campos2025gaea}, and an important testbed for open-world multimodal reasoning and knowledge-grounded intelligence~\cite{geng2026geobrowse}.

Traditional geo-localization research has focused on visual perception through classification approaches~\cite{weyand2016planet,translocator} and feature retrieval~\cite{vivanco2023geoclip,haas2024pigeon}. However, their effectiveness largely depends on large-scale, predefined, and closed-world reference datasets. With the advancement of deep learning, recent work has begun to leverage large language models equipped with world knowledge to reason over visual cues. By enhancing visual reasoning abilities through supervised fine-tuning~\cite{li2024georeasoner,wang2025gre} or reinforcement learning~\cite{wang2025geovista,jia2026spotagent,liu2026thinking}, geo-localization is gradually shifting from pure pattern recognition toward perception-driven reasoning.

\begin{figure}[t]
    \centering
    \includegraphics[width=\linewidth]{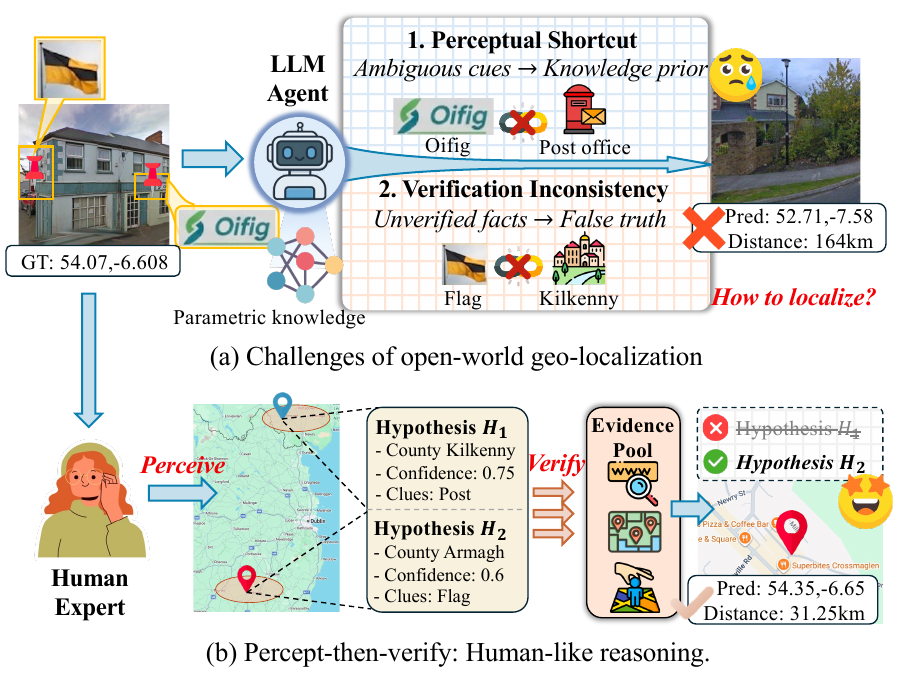}
    \caption{Challenges of geo-localization.}
    \label{fig:introduction}
\end{figure}

Though promising, recent LVLM post-training paradigms~\cite{wang2025geovista,recognition-globe} focus on extending reasoning depth, yet they still struggle to maintain reliable reasoning when perceptual evidence is ambiguous and intermediate reasoning steps lack reliable auditing. These limitations reveal two core challenges (see Figure~\ref{fig:introduction}) inherent to perception-driven reasoning:

\textbf{1) Perceptual Shortcut:} When visual evidence is partial or ambiguous, LVLMs tend to map salient local cues to high-frequency geographic regions based on blurred text or architectural styles. Although test-time scaling techniques, such as parallel rollouts~\cite{amap}, expand the search space, they increase inference costs without fundamentally resolving the inherent unreliability of perception-driven reasoning in single pass~\cite{tran2026single}.

\textbf{2) Verification Inconsistency:} Existing geo-localization research largely overlooks the importance of evidence-grounded verification. While agentic RL~\cite{urbanagent} has demonstrated the utility of external tools, unconstrained tool invocation in open environments may introduce lengthy contexts that mix useful evidence with irrelevant retrieval results, undermining the consistency between initial visual cues and final conclusions. For example, when a blurred storefront retrieves similar POIs from multiple cities, the model may follow an unverified search result and converge to a plausible but visually inconsistent location.

To tackle these issues, we revisit the intrinsic nature of ambiguous reasoning in open-world geo-localization. We argue that the key lies in \textit{simulating human-like cognitive behavior}. In geographic discovery games such as GeoGuessr~\cite{talreja2026georc,cheng2025geoguess-multimodal}, expert players rarely rely on a single-shot intuition, but instead operate in a continuous \textbf{perceive-then-verify} loop. They first perceive multi-granularity visual cues to form and mentally rank a set of plausible hypotheses, continuously cross-checking these hypotheses against external geographic and map-based context to confirm, reject, or refine their beliefs. 

Inspired by the human-like reasoning process, we first reformulate geo-localization as a perception-verification paradigm, and then propose \textbf{GeoPAVE}, a bi-level agentic framework that decomposes geo-localization into human-like perception and verification stages. Our GeoPAVE consists of 1) an \textit{Agentic Perception Module} that generates uncertainty-aware hypotheses within a single reasoning pass, while leveraging uncertainty quantification to determine verification actions; and 2) a \textit{Verification Module} that grounds candidate hypotheses through strategic tool invocation, enabling evidence-based hypothesis refinement. By adaptively adjusting hypotheses with verifiable external evidence, GeoPAVE effectively suppresses hallucinations while preserving contextual consistency. To facilitate rigorous evaluation, we construct \textbf{PAVED}, a high-quality dataset sourced from human check-ins, containing rich tool-use evidence-based hypothesis verification reasoning trajectories generated from GeoPAVE. Our contributions can be summarized as follows:
 
\begin{itemize}[leftmargin=*, nosep]
\item We revisit open-world geo-localization based on human cognitive reasoning, reformulating it as a \textbf{perceive-then-verify} paradigm and proposing a perception--verification agentic reasoning framework \textbf{GeoPAVE}. This framework naturally produces high-quality reasoning trajectories with rich evidence-based verification.

\item We introduce an agentic perception module that condenses candidate-space exploration into a single reasoning pass with uncertainty estimation, together with a structured verification module that maps uncertainty signals to specialized tool-based verification skills, performing tool-augmented evidence retrieval and iterative hypothesis refinement.

\item Extensive experiments across multiple benchmarks demonstrate that GeoPAVE achieves strong and competitive performance against representative retrieval-based, agentic, and closed-source LVLM baselines. In addition, we present PAVED (Perception-and-Verification-Engine-Dataset), comprising streetview images collected from human check-ins, paired with verification procedures and tool-call trajectories.

\end{itemize}

\begin{figure*}[t]
    \centering
    \includegraphics[width=\textwidth]{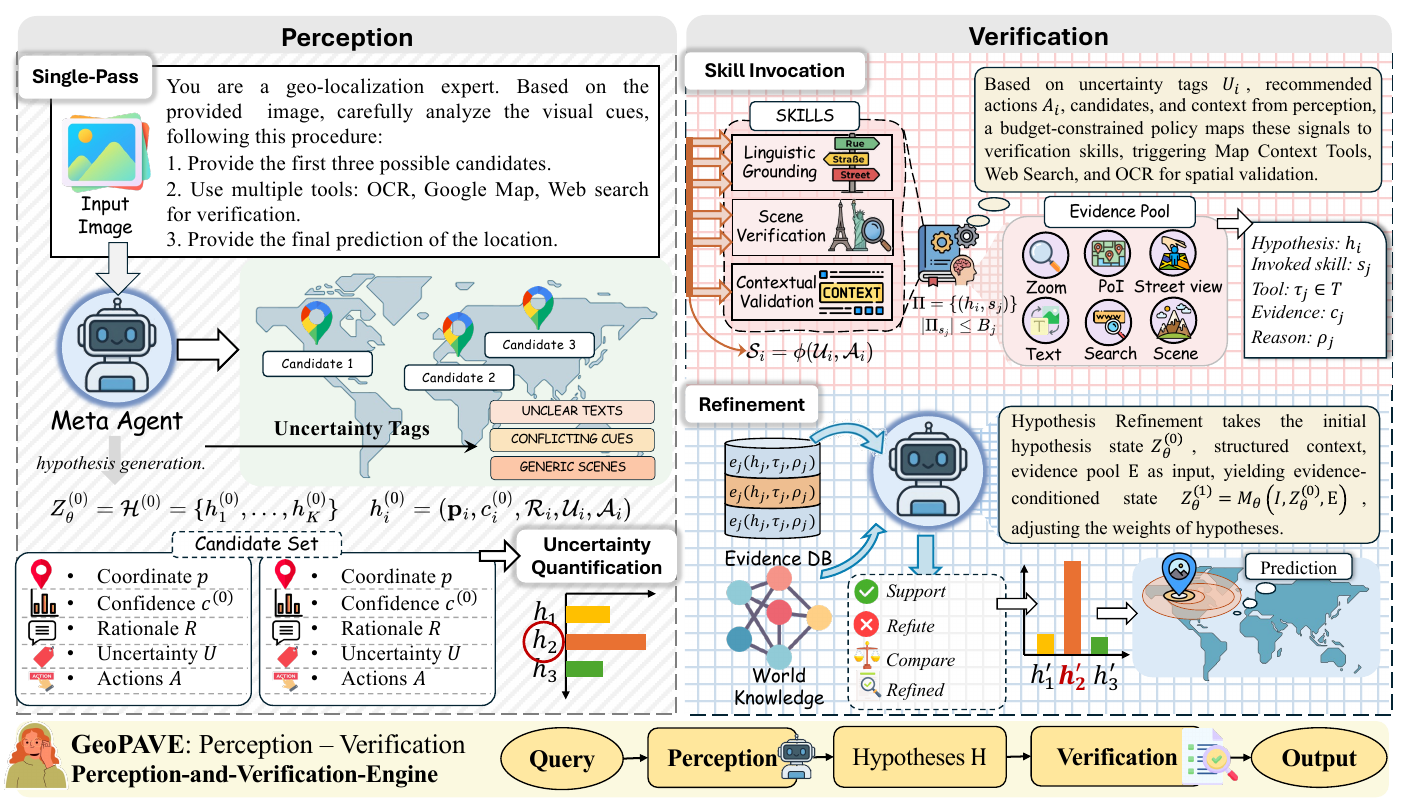}
    \caption{Overview of GeoPAVE.}
    \label{fig:framework}
\end{figure*}

\section{Preliminaries}

\subsection{Problem Statement}
  
Given an image $\mathcal{I}$, geo-localization aims to predict its coordinate $\hat{\mathbf{p}}=(\text{lat},\text{lon})$. We formulate it as a \textbf{perception-verification problem}: a vision-language model first produces candidate hypotheses $\mathcal{H}=\{h_i\}_{i=1}^{k}$ with confidence scores and uncertainty tags, while external tools retrieve evidence $\mathcal{E}=\{\mathbf{e}_j\}_{j=1}^{m}$ to refine them. The final coordinate is selected by evidence-conditioned scoring over the refined hypotheses.

\subsection{Related Works}

\paragraph{Open-World Geo-Localization.} 
Open-World geo-localization aims to predict the precise geographic location of an image across the globe. Traditional computer vision methods tackle this through classification by discretizing the Earth into predefined cells~\cite{seo2018cplanet,translocator,ghasemi2025geotoken} or retrieval by matching queries against large-scale visual databases~\cite{shatwell2025gt-loc,haas2024pigeon,jia2024g3}. However, these closed-world approaches heavily rely on massive datasets and struggle to generalize to open-world conditions~\cite{jiang2026cost}. Recently, large vision-language models (LVLMs) have addressed this limitation by leveraging multimodal reasoning and extensive world knowledge to directly infer locations~\cite{li2024georeasoner,Geo-ADAPT,recognition-globe,li2026correcting}. While their explicit reasoning improves interpretability, the resulting reasoning chains are often long and complex, which can easily lead to hallucinations and thus undermine reliability. Moreover, the absence of verification mechanisms further limits the trustworthiness and precision of their final predictions.

\paragraph{Agentic Reasoning System.} Recent progress in large language models (LLMs) has been driven by integrating external tools, enabling agentic multimodal systems to ground perception through environment interactions and extend beyond parametric knowledge~\cite{schick2023toolformer,khaki2026vistira,qin2024toolllm}. To optimize tool use and long-horizon decision-making, further studies explore agentic reinforcement learning~\cite{li2026adacurl,qin2025learn} and test-time scaling via parallel or multi-agent exploration~\cite{zhu2025scaling,qiao2025webresearcher,guo2026agentsense}. In the geo-localization domain, researchers employ multi-agent debate~\cite{han2025swarm,zheng2025graphgeo} and agentic RL~\cite{amap,jia2026spotagent} to improve reasoning depth. However, these methods demand expensive computational resources, as their performance gains heavily rely on increased reasoning trajectories. In contrast, our proposed \textbf{GeoPAVE} achieves higher reliability via a structured bi-level perception-verification stage. 
  
\section{Methodology}

\subsection{Overview}

GeoPAVE constructs a bi-level verification architecture to address the challenges of ambiguous visual reasoning in geo-localization, which decouples hypothesis generation from evidence-based validation. GeoPAVE grounds predictions in verifiable external evidence through structured tool invocation, providing a \textbf{hypothesis-verification} process: a perception module that performs single-pass perception to explore the hypothesis space, and a verification module that implements non-parametric refinement through cross-modal evidence aggregation from heterogeneous knowledge sources~\cite{zhong2025multimodal}. The pipeline is shown in Figure~\ref{fig:framework}.

\subsection{Perception Module}

\paragraph{Single-Pass Candidate Generation.}
For input image $\mathcal{I}$, the perception agent generates a first-pass hypothesis state $Z_\theta^{(0)}=\mathcal{H}^{(0)}=\{h_1^{(0)}, \ldots, h_K^{(0)}\}$, where $K \in \{1, 2, 3\}$. This process produces multiple potential outcomes in a single inference pass. Each structured hypothesis $h_i^{(0)} = (\mathbf{p}_i, c_i^{(0)}, \mathcal{R}_i, \mathcal{U}_i, \mathcal{A}_i)$ consists of a predicted coordinate $\mathbf{p}_i = (\text{lat}_i, \text{lon}_i)$, an initial confidence score $c_i^{(0)} \in [0,1]$, a natural language rationale $\mathcal{R}_i$, uncertainty tags $\mathcal{U}_i \subseteq \mathcal{U}$ from a predefined vocabulary, and verification actions and tools $\mathcal{A}_i$. 

To enable reliable test-time scaling, we model uncertainty as structured tags that guide hypothesis verification. Given an initial hypothesis confidence $c_i^{(0)}$, uncertainty signals $\mathcal{U}_i$ derived from conflicting cues trigger confidence calibration $c_i^{(0)} \leftarrow f(c_i^{(0)}, \mathcal{U}_i)$ suppressing overconfident predictions before verification.

\paragraph{Uncertainty Quantification.} 

GeoPAVE represents uncertainty as explicit routing signals for tool-call process. If visual cues, OCR-like text snippets, or candidate semantics conflict, the perception module lowers the adjusted confidence and records the unresolved ambiguity through fields such as \texttt{CONFLICTING\_CUES} and uncertainty tags $\mathcal{U}$. This structured context describes image-internal observations and verification needs, which indicates that visual features and textual semantics diverge) to mitigate overconfident predictions on ambiguous data (detailed information in Appendix~\ref{sec:appendix_verification}). The tag set activates verification skills through a mapping $\phi:{\mathcal{U}}\rightarrow{\mathcal{S}}$ in the verification stage, without skill-based learning policies.

\subsection{Verification Module}
\paragraph{Conditional Skill Invocation.}
GeoPAVE formulates verification as conditional skill invocation, mapping epistemic signals to verification operations. Specifically, a verification skill $s \in \mathcal{S}$ is defined as a tuple $s = (\mathcal{U}_s, t_s, \psi_s)$, where $\mathcal{U}_s$ denotes uncertainty tags that activate the skill, $t_s \in \mathcal{T}$ denotes the external knowledge source, and $\psi_s: \mathcal{E} \times \mathcal{H} \rightarrow \mathbb{R}$ quantifies the mutual information $I(\mathbf{p}; \mathbf{e} \mid \mathcal{I}, \mathcal{E}_t)$ between the evidence item $\mathbf{e}$ and the true location $\mathbf{p}$ conditioned on the image and prior evidence, where $\mathcal{H}$ denotes the hypothesis set and $\mathcal{E}$ denotes the evidence pool. GeoPAVE instantiates three core skills: 

\begin{itemize}[leftmargin=*, nosep]
    \item[{1)}] \textit{Linguistic Grounding} extracts script and language features from street signs to produce geographic constraints.
    \item[{2)}] \textit{Scene Verification} queries search engines with visual scenes to retrieve metadata.
    \item[{3)}] \textit{Contextual Validation} checks candidate locations against elevation, waterbody, topographic, and nearby-POI evidence from geographic databases to assess spatial consistency.
\end{itemize}

The skill selection policy $\pi: \mathbf{s}_t \rightarrow \mathcal{S}$ is rule-based, mapping uncertainty tags to verification skills and thereby enabling adaptive verification without policy learning. Each invoked skill produces a structured evidence record
$\mathbf{e}_j = (i_j, \tau_j, o_j, y_j, \rho_j, \Delta_j, v_j)$,
where $i_j$ indexes the target hypothesis in $\mathcal{H}$, $\tau_j \in \mathcal{T}$ denotes the evidence source, $o_j$ is the tool observation or retrieved content, $\rho_j$ provides a compact reason code, $y_j \in \{\textsc{Support}, \textsc{Refute}, \textsc{Refine}\}$ denotes the verification decision, $\Delta_j$ is the rule-based score update induced by this evidence, and $v_j \in \{0,1\}$ indicates whether the evidence triggers a hard veto. The outputs of all invoked skills form the evidence pool $\mathcal{E}=\{\mathbf{e}_j\}_{j=1}^{m}$, which serves as the grounded interface between tool execution and hypothesis refinement.

Specifically, \textsc{Support} increases the confidence of a hypothesis when the retrieved evidence is consistent with the candidate location, \textsc{Refute} decreases the confidence or rejects a hypothesis when severe geographic contradictions are observed, and \textsc{Refine} updates the candidate location or its supporting context when newly discovered textual or spatial evidence can be successfully grounded.

\begin{figure}[t]
    \centering
    \includegraphics[width=\linewidth]{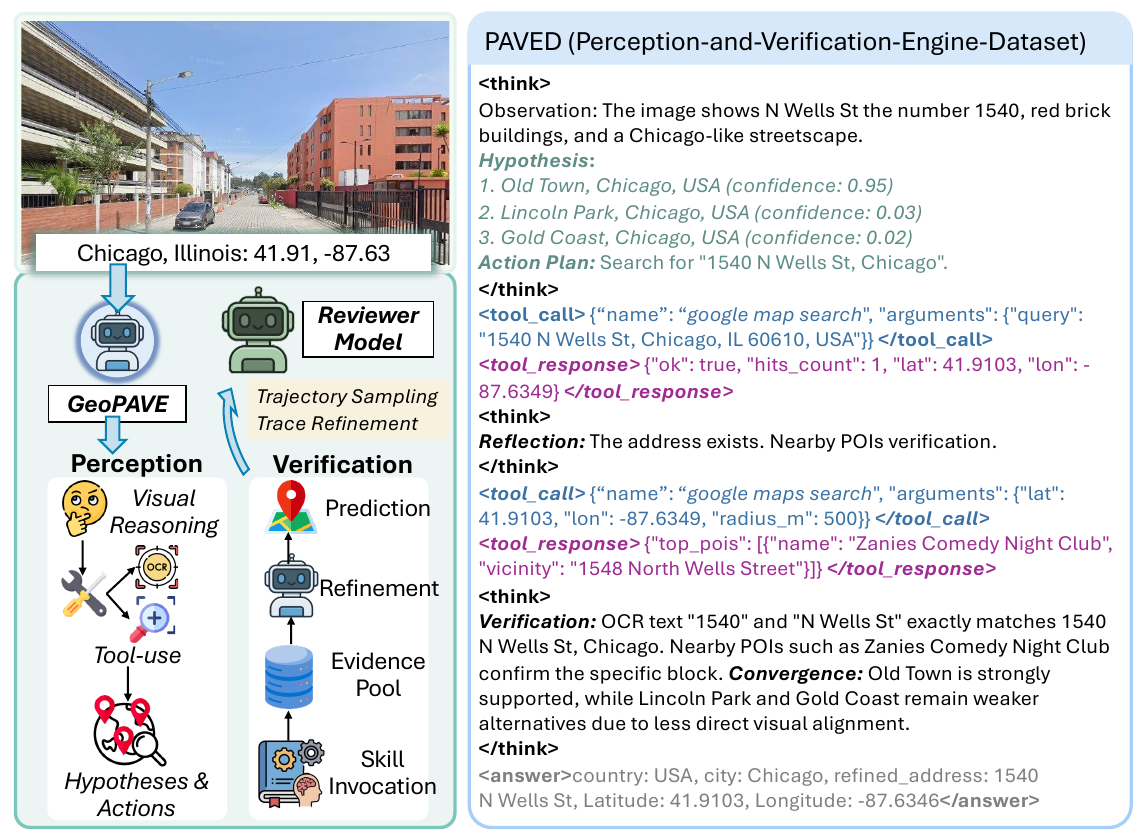}
    \caption{PAVED trace construction pipeline. }
    \label{fig:cot}
\end{figure}

\paragraph{Hypothesis Refinement.}
After evidence collection, GeoPAVE performs a second reasoning pass to reassess the initial hypotheses under the grounded evidence pool $\mathcal{E}$. Specifically, it takes the initial hypothesis state $Z_\theta^{(0)}=\{(h_i^{(0)},c_i^{(0)})\}_{i=1}^{K}$, together with the image $\mathcal{I}$ and the evidence pool $\mathcal{E}$, as input to the refinement agent $\mathcal{M}_\theta$, as defined in Eq.~\ref{eq:refined_hypothesis}. The agent re-evaluates each candidate hypothesis according to supportive, refuting, and refinement evidence, producing an updated hypothesis state $Z_\theta^{(1)}$. In this process, supportive evidence increases the candidate confidence, refuting evidence decreases it, and severe contradictions trigger a veto that excludes the candidate from final selection.

\begin{equation}
    Z_\theta^{(1)} = \mathcal{M}_\theta(\mathcal{I}, Z_\theta^{(0)}, \mathcal{E}) = \{(h_i^{(1)}, c_i^{(1)})\}_{i=1}^{K}.
    \label{eq:refined_hypothesis}
\end{equation}

The final hypothesis is then deterministically selected as the highest-scoring valid candidate: $\hat h = \arg\max_{h_i\in Z_\theta^{(1)}} c_i^{(1)}$. By anchoring this second reasoning pass explicitly to auditable evidence, the model effectively reduces context drift and preserves reasoning consistency in ambiguous, long-horizon visual scenarios.

\section{Datasets}


\newcommand{\cmark}{\ding{51}}
\newcommand{\xmark}{\ding{55}}
\newcommand{\pmark}{$\circ$}

\begin{table}[t]
\centering
\small
\renewcommand{\arraystretch}{1.16} 
\setlength{\tabcolsep}{3.6pt}      

\begin{tabularx}{\columnwidth}{l *{7}{>{\centering\arraybackslash}X}}
\toprule
\textbf{Dataset} &
\textbf{GC} &
\textbf{RL} &
\textbf{HR} &
\textbf{DV} &
\textbf{NE} &
\textbf{CoT} &
\textbf{Tool} \\
\midrule
Im2GPS3k          & \cmark & \xmark & \xmark & \xmark & \cmark & \xmark & \xmark \\
MAPBench          & \cmark & \pmark & \pmark & \pmark & \cmark & \pmark & \cmark \\
EarthWhere        & \cmark & \pmark & \pmark & \pmark & \cmark & \pmark & \pmark \\
GeoRC             & \pmark & \cmark & \pmark & \xmark & \pmark & \cmark & \xmark \\
GeoBench          & \cmark & \cmark & \cmark & \cmark & \cmark & \pmark & \cmark \\
\midrule
\textbf{PAVED} &
\textbf{\cmark} &
\textbf{\cmark} &
\textbf{\cmark} &
\textbf{\cmark} &
\textbf{\cmark} &
\textbf{\cmark} &
\textbf{\cmark} \\
\bottomrule
\end{tabularx}

\begin{flushleft}
{\footnotesize
GC: global coverage; 
RL: reasonable localizability; 
HR: high-resolution imagery; 
DV: data variety; 
NE: nuanced evaluation; 
CoT: paired reasoning trajectory; 
Tool: explicit tool-call trajectory. 
\cmark: yes, \pmark: partial, \xmark: no.}
\end{flushleft}

\caption{Comparison of geolocalization benchmarks. PAVED combines globally distributed human check-in imagery, localizability filtering, high-resolution visual evidence, distance- and address-level evaluation, and auditable tool-grounded reasoning trajectories.}
\label{tab:benchmark}
\end{table}
  
\subsection{Data Collection.}

We construct \textbf{PAVED} (Perception-and-Verification-Engine-Dataset) from the Foursquare check-in corpus~\cite{foursquare}, containing 2,524 manually filtered street-view images collected from worldwide human check-ins. PAVED focuses on accessible places with clear semantic cues, such as commercial areas, public streets, and residential districts, better reflecting realistic open-world geo-localization scenarios. As summarized in Table~\ref{tab:benchmark}, PAVED complements existing benchmarks by jointly supporting global coverage, geo-localizability, high-resolution imagery, nuanced evaluation, and paired tool-grounded reasoning trajectories. Additional details on geographic distribution, coverage, and sampling are provided in the Appendix~\ref{sec:appendix_PAVED}.

\subsection{Chain-of-Thought}
  
\begin{table*}[htbp]
\centering
\captionsetup{font=small}
\renewcommand{\arraystretch}{1.4}
\setlength{\tabcolsep}{1.5pt}
\definecolor{darkgraytext}{HTML}{4A4A4A}
\newcommand{\gaincell}[1]{\textcolor{darkgraytext}{\footnotesize\textit{#1}}}

\resizebox{\textwidth}{!}{%
\normalsize
\begin{tabular}{l ccccc ccccc ccccc}
\toprule
\multirow{2}{*}{\textbf{Model}} &
\multicolumn{5}{c}{\textbf{PAVED}} &
\multicolumn{5}{c}{\textbf{MAPBench-V2}} &
\multicolumn{5}{c}{\textbf{IM2GPS3K}} \\
\cmidrule(lr){2-6}
\cmidrule(lr){7-11}
\cmidrule(lr){12-16}

& 500m & 2km & 25km & 200km & 750km
& 500m & 2km & 25km & 200km & 750km
& 500m & 2km & 25km & 200km & 750km \\
\midrule

\rowcolor{gray!10}
GeoCLIP
& 2.58 & 7.49 & 24.52 & 37.08 & 59.31
& 0.88 & 1.59 & 6.01 & 12.39 & 32.29
& 8.94 & 18.35 & 32.17 & 49.45 & 66.53 \\

GeoReasoner
& 0.99 & 5.23 & 31.42 & 36.57 & 50.71
& 0.00 & 1.46 & 11.76 & 21.44 & 37.09
& 2.04 & 10.98 & 25.19 & 36.24 & 52.82 \\

\rowcolor{gray!10}
GAEA
& 1.86 & 7.25 & 34.39 & 41.64 & 61.69
& 0.42 & 5.26 & 25.45 & 34.84 & 56.24
& -- & -- & 32.61 & 51.42 & -- \\

GeoVista
& 1.55 & 6.74 & 39.22 & 44.97 & 63.31
& 0.25 & 4.17 & 22.86 & 27.36 & 33.54
& 3.47 & 13.55 & 28.06 & 35.90 & 44.61 \\

\rowcolor{gray!10}
GLOBE
& 0.67 & 2.85 & 14.86 & 20.80 & 38.99
& 1.20 & 3.91 & 13.25 & 42.84 & 62.53
& 2.10 & 8.91 & 35.12 & 50.24 & 64.61 \\

Thinking~with~Map~1 rollout
& 3.69 & 9.65 & 40.53 & 47.38 & 70.55
& 13.73 & 23.20 & 49.52 & 57.95 & 76.60
& 7.39 & 19.81 & 41.09 & 56.45 & 74.62 \\

\rowcolor{gray!10}
Thinking~with~Map~4 rollouts
& 9.09 & 17.89 & 51.03 & 58.07 & 78.30
& \textbf{22.15} & \textbf{27.08} & 40.26 & 53.43 & 62.48
& 7.94 & 16.52 & 29.70 & 38.44 & 48.18 \\

Qwen3-VL-235B
& 2.42 & 8.80 & 26.90 & 45.64 & 65.41
& 2.88 & 10.81 & 34.00 & 44.85 & 62.54
& 6.94 & 18.08 & 36.54 & 52.42 & 70.24 \\

\rowcolor{gray!10}
Gemini-3.1-Pro
& 11.10 & 17.00 & \underline{57.50} & 63.40 & 77.20
& 12.37 & 22.51 & \underline{53.48} & \textbf{76.22}
& \underline{82.72}
& -- & -- & -- & -- & -- \\

SmileGeo
& -- & -- & -- & -- & --
& -- & -- & -- & -- & --
& -- & -- & 47.77 & -- & -- \\

\rowcolor{gray!10}
GPT-5.5
& 5.31 & 13.16 & 39.42 & 47.28 & 67.51
& 7.68 & 18.48 & 34.63 & 43.68 & 58.70
& \textbf{19.19} & \textbf{35.37} & \textbf{56.52}
& \textbf{69.44} & \textbf{81.38} \\

Gemini-3.0-Flash
& \underline{14.92} & \underline{23.34} & 54.74
& \underline{64.45} & \underline{78.80}
& 11.14 & 20.57 & 45.14 & 53.61 & 65.96
& 15.42 & 29.70 & 51.16 & 64.17 & 80.08 \\

\rowcolor[HTML]{E6F0FF}
GeoPAVE (Ours)
& \textbf{16.87} & \textbf{26.89} & \textbf{59.96}
& \textbf{69.35} & \textbf{85.86}
& \underline{13.81} & \underline{24.83} & \textbf{63.19}
& \underline{70.42} & \textbf{86.87}
& \underline{15.92} & \underline{30.51} & \underline{52.30}
& \underline{66.18} & \underline{81.24} \\

\rowcolor[HTML]{F2F7FF}
\gaincell{Gain (\%)}
& \gaincell{+1.95} & \gaincell{+3.55} & \gaincell{+5.22}
& \gaincell{+4.90} & \gaincell{+7.06}
& \gaincell{+2.67} & \gaincell{+4.26} & \gaincell{+18.05}
& \gaincell{+16.81} & \gaincell{+20.91}
& \gaincell{+0.50} & \gaincell{+0.81} & \gaincell{+1.14}
& \gaincell{+2.01} & \gaincell{+1.16} \\

\bottomrule
\end{tabular}%
}

\caption{Comparison of geolocation accuracy on PAVED, MAPBench-V2,
and IM2GPS3K under different error thresholds (values in \%).
\textbf{Bold} and \underline{underlined} values denote the best and
second-best reported results, respectively. GeoPAVE uses
Gemini-3.0-Flash as base model. The Gain row reports the
absolute improvement of GeoPAVE over Gemini-3.0-Flash in percentage
points (pp).}
\label{tab:main_results}
\end{table*}

\paragraph{Structured CoT.}
We combine PAVED with agentic perception-verification reasoning. Unlike conventional datasets that merely provide final coordinates and perception-driven reasoning traces, \textbf{PAVED} features structured Chain-of-Thought (CoT) trajectories that explicitly decouple reasoning into hypothesis generation, evidence verification, and candidates refinement. Given an image $\mathcal{I}$ and its ground-truth coordinate $\mathbf{p}^\ast$, we formulate an answer-conditioned expert trajectory as Eq.~\ref{eq:gt_cot}:
\begin{equation}
\label{eq:gt_cot}
\Gamma^\ast = \{ Z^{(0)}, \mathcal{E}, Z^{(1)}, \mathbf{p}^\ast \}, 
\end{equation}
where the initial hypothesis state $Z^{(0)} = \{h_i^{(0)}\}_{i=1}^{K}$ comprises candidates formalized as $h_i^{(0)} = (\mathbf{p}_i, c_i^{(0)}, \mathcal{R}_i, \mathcal{U}_i, \mathcal{A}_i)$, capturing the coordinate, initial confidence, perception rationale, uncertainty, and recommended verification actions, respectively. The trajectory progresses through $\mathcal{E}$, representing the sequence of tool actions and observations, ultimately yielding the refined hypothesis state $Z^{(1)}$.

\paragraph{Trajectory Curation.}
The generation follows a rigorous two-stage curation workflow: 1) The base GeoPAVE framework generates a preliminary structured trace that logs candidate generation, tool interactions, evidence evaluations, and the final convergence state; 2) a stronger reviewer model is introduced to audit the initial trace utilizing rejection sampling to enforce tool-call validity, evidence consistency, and prediction precision ($d(\hat{\mathbf{p}}, \mathbf{p}^\ast) < 5\text{km}$), as shown in Figure~\ref{fig:cot}. By formatting the retained trajectories with structured tags (e.g., \texttt{<think>}) to separate internal reasoning from external observations, we provide a high-quality dataset with transparent reasoning traces, offering a valuable resource for analyzing verifiable agentic geo-localization.

\section{Experiments}

To examine the effectiveness of GeoPAVE and bi-level architecture, we organize our experiments around three research questions: 
\begin{itemize}[leftmargin=*]
    \item \textbf{RQ1:} Can GeoPAVE achieve fine-grained geo-localization in ambiguous visual geo-localization under lower cost?
    \item \textbf{RQ2:} How do the individual components of GeoPAVE's bi-level architecture contribute to the overall prediction accuracy? 
    \item \textbf{RQ3:} Does hypothesis-verification effectively mitigate \textit{perceptual shortcuts} and prevent \textit{contextual inconsistency} during long-horizon ambiguous visual reasoning of geo-localization?
\end{itemize}

\subsection{Experiment Setup}

\paragraph{Benchmark.}
We evaluate GeoPAVE on three complementary benchmarks. \textbf{PAVED} is our 2{,}524-image benchmark constructed from human check-ins, covering worldwide street-view images with visual cues suitable for evidence-based geographic reasoning. \textbf{IM2GPS3K}~\cite{im2gps3k} is a standard public global geo-localization benchmark containing 2{,}997 Flickr-sourced geotagged images. \textbf{MAPBench-V2}~\cite{amap} is a China-centered benchmark covering urban POI and street-view scenes across multiple cities, and we use its 2.4K-image test split for evaluation. Detailed descriptions of these public datasets are provided in the appendix~\ref{sec:appendix_datasets}. Together, these datasets test global open-world geo-localization, legacy Internet-photo geo-localization, and China-specific fine-grained urban geo-localization.

\paragraph{Baselines.}
We compare GeoPAVE with representative retrieval-based, fine-tuned, reasoning-based, and tool-augmented geo-localization systems. The retrieval baseline is GeoCLIP~\cite{vivanco2023geoclip}. Fine-tuned or geolocation-specialized LVLM baselines include GAEA~\cite{campos2025gaea}, GeoReasoner~\cite{li2024georeasoner}, and GLOBE~\cite{recognition-globe}. Agentic and web-augmented baselines include GeoVista~\cite{wang2025geovista} and Thinking with Map~\cite{amap}. We also consider strong closed-source LVLMs such as GPT-5.5 and Gemini-3.1-Pro. To ensure a fair comparison with Thinking with Map~\cite{amap}, we consider both its original single-rollout setting and test-time scaling approaches as baselines. Specifically, we evaluate the 4-rollout setting by incorporating Gemini-3.0-Flash  to provide a comprehensive assessment. More details are in Appendix~\ref{sec:appendix_Baselines}.

\paragraph{Metrics.}

We evaluate geo-localization quality from final-coordinate accuracy and verification-trace quality. For coordinate accuracy, we report distance-threshold accuracy $\mathrm{Acc@}r$, where a prediction is correct if its haversine distance to the ground-truth coordinate is within $r$. $r$ is set as $500\mathrm{m}$, $2\mathrm{km}$, $25\mathrm{km}$, $200\mathrm{km}$, and $750\mathrm{km}$ respectively, covering street-level to regional localization.

Beyond final accuracy, we introduce three trace-based metrics to examine candidate diversity and evidence-grounded correction. Let $p_i^\ast$ denote the ground-truth coordinate of the $i$-th image, $\mathcal{H}_i^{(0)}=\{p_{ij}\}_{j=1}^{K_i}$ denote the initial candidate coordinates, $\tilde{p}_i$ denote the initial prediction, and $\hat{p}_i$ denote the final prediction.

\noindent
\textbf{Candidate Oracle Recall (COR@$r$).}
COR@$r$ measures whether the initial hypothesis set covers the correct geographic region:
\begin{equation}
\mathrm{COR}@r = \frac{1}{N}\sum_{i=1}^{N}
\mathbf{1}\left[\min_j d(p_{ij},p_i^\ast)\le r\right].
\end{equation}

\noindent
\textbf{Correction Recall (CR@$r$).}
CR@$r$ evaluates whether verification recovers initially failed samples. Given $\mathcal{S}_{\tau}=\{i\mid e_i^{(0)}>\tau\}$, where $e_i^{(0)}=d(\tilde{p}_i,p_i^\ast)$, we compute:
\begin{equation}
\mathrm{CR}@r \mid \tau =
\frac{1}{|\mathcal{S}_{\tau}|}
\sum_{i\in\mathcal{S}_{\tau}}
\mathbf{1}\left[d(\hat{p}_i,p_i^\ast)\le r\right].
\end{equation}

\noindent
\textbf{Log Correction Gain (LCG).}
LCG measures the logarithmic error reduction from the initial prediction to the final prediction:
\begin{equation}
\mathrm{LCG}=
\frac{1}{N}\sum_{i=1}^{N}
\log_{10}\frac{d(\tilde{p}_i,p_i^\ast)+1}{d(\hat{p}_i,p_i^\ast)+1}.
\end{equation}

\paragraph{Settings.}
For closed-source models, we perform inference using their official APIs, while open-source models are evaluated on NVIDIA A6000 GPUs. All models use the same Nominatim map-service interface to ensure a fair comparison under a shared geocoding backend. GeoPAVE is primarily instantiated with Gemini-3.0-Flash, while GPT-4o-mini is additionally used as a weaker backbone in the robustness study. Further implementation details are provided in Appendix~\ref{sec:appendix_Metrics}.


\begin{table}[t]
\centering
\small
\setlength{\tabcolsep}{3.8pt}
\begin{tabular}{@{}lccccc@{}}
\toprule
\multicolumn{1}{c}{\textbf{Model}}
& \multicolumn{5}{c}{\textbf{Accuracy (\%)}} \\
\cmidrule(lr){2-6}
& \textbf{500\,m}
& \textbf{2\,km}
& \textbf{25\,km}
& \textbf{200\,km}
& \textbf{750\,km} \\
\midrule
P-Only
& 12.72 & 24.29 & 56.63 & 66.25 & 85.38 \\

V-Only
& 14.22 & 22.28 & 55.84 & 65.43 & 84.61 \\

V w/o OCR
& 15.16 & 24.31 & 58.91 & 68.54 & 85.32 \\

V w/o Web
& 11.87 & 22.86 & 56.43 & 69.00 & 85.06 \\

V w/o Map
& 13.58 & 25.36 & 57.52 & 67.82 & 84.31 \\

\midrule
GeoPAVE (${\scriptstyle K=4}$)
& \textbf{16.87}
& \textbf{26.89}
& 59.96
& 69.35
& \textbf{85.86} \\

GeoPAVE (${\scriptstyle K=3}$)
& 15.04
& 25.06
& \textbf{61.84}
& \textbf{70.25}
& 85.53 \\
\bottomrule
\end{tabular}

\caption{
Ablation study of different GeoPAVE configurations and
maximum hypothesis budgets on PAVED.
P and V denote the Perception and Verification modules,
respectively. The default setting uses $K_{\max}=3$.
}
\label{tab:ablation_results}
\end{table}

\subsection{Overall Performance (RQ1)}
We compare GeoPAVE with representative retrieval-based, fine-tuned, agentic, and closed-source LVLM baselines on PAVED, MAPBench-V2, and IM2GPS3K, as summarized in Table~\ref{tab:main_results}. On PAVED, GeoPAVE achieves the best performance across all thresholds. Compared with Gemini-3.1-Pro, a strong closed-source LVLM baseline, GeoPAVE improves Acc@500\,m from 11.10\% to 16.87\% and Acc@2\,km from 17.00\% to 26.89\%. The consistent gains from fine-grained to coarse-grained thresholds indicate that the perceive-then-verify design improves fine-grained geo-localization, which is especially important for PAVED, where many samples require text cues, POI, and contextual cues rather than relying on a single visual prior.

The remaining results highlight where GeoPAVE is most effective. The stronger performance of Thinking~with~Map 4 rollouts on MapBenchV2 can be attributed to the large-scale agentic RL training on such data, along with test-time rollouts enlarging the search space for reasoning. However, GeoPAVE outperforms it from Acc@25km onward, suggesting that while our method is slightly weaker in exact fine-grained geo-localization, it remains effective at coarse-grained scenarios. For the Flickr-based benchmark, GeoPAVE slightly trails GPT-5.5, which is expected since many images are indoor scenarios, which favor larger closed-source LVLMs with stronger general visual understanding and broader world knowledge. These results suggest that GeoPAVE is most effective when visual observations can be connected to verifiable local evidence, while remaining competitive under less constrained image distributions.

\subsection{Robustness Study (RQ1)}
\begin{table}[t]
\centering
\captionsetup{font=small}
\renewcommand{\arraystretch}{1.15}
\setlength{\tabcolsep}{3.5pt}

\resizebox{\columnwidth}{!}{%
\small
\begin{tabular}{ll ccc}
\toprule
\textbf{Dataset} &
\textbf{Radius} &
\shortstack{\textbf{GPT-4o-mini}} &
\shortstack{\textbf{GeoPAVE}} &
\textbf{Gain (\%)} \\
\midrule

\multirow{6}{*}{PAVED}
& 500m  & 2.77  & 2.87  & \textit{+0.10} \\
& 2km   & 10.62 & 11.36 & \textit{+0.74} \\
& 25km  & 34.90 & 36.84 & \textit{+1.94} \\
& 100km & 38.11 & 42.30 & \textit{+4.19} \\
& 200km & 40.77 & 45.05 & \textit{+4.28} \\
& 750km & 52.93 & 62.28 & \textit{+9.35} \\
\midrule

\multirow{6}{*}{\makecell{MAP\\Bench\\V2}}
& 500m  & 0.375 & 0.872 & \textit{+0.497} \\
& 2km   & 3.13  & 4.31  & \textit{+1.18} \\
& 25km  & 11.64 & 14.72 & \textit{+3.08} \\
& 100km & 12.93 & 16.87 & \textit{+3.94} \\
& 200km & 15.06 & 20.94 & \textit{+5.88} \\
& 750km & 29.16 & 35.84 & \textit{+6.68} \\

\bottomrule
\end{tabular}%
}

\caption{Reliability of GeoPAVE on PAVED and MAPBench-V2 with a weaker
backbone. GPT-4o-mini is evaluated standalone and within
GeoPAVE, with gains reported in percentage points.}
\label{tab:gpt4omini_reliability}
\end{table}

To examine whether GeoPAVE's gains depend on a specific strong backbone, we replace Gemini-3.0-Flash with the weaker GPT-4o-mini for both base reasoning and verification. As shown in Table~\ref{tab:gpt4omini_reliability}, GeoPAVE consistently improves the corresponding API baseline across all thresholds on both PAVED and MAPBench-V2, with gains increasing to 9.35\% on PAVED and 6.68\% on MAPBench-V2 at 750 km. These results suggest that the perception-verification architecture remains effective under a weaker backbone, instead of relying on specific models. The widening gains at larger radii further indicate that evidence-grounded verification is effective at preventing catastrophic geographic errors when backbone capacity is limited.

\subsection{Ablation Study (RQ2)}

\paragraph{Effect of perception and verification.}
We first evaluate the contribution of GeoPAVE's two-stage design. As shown in Table~\ref{tab:ablation_results}, both \textit{P-only} and \textit{V-only} underperform the full model. Removing verification reduces Acc@500m from 16.87\% to 12.72\%, showing that initial hypotheses still require evidence-based refinement. Verification-only also degrades performance, indicating that verification is less effective without uncertainty-aware candidate generation. The distinct degradation patterns further suggest that perception preserves broad candidate coverage, whereas verification is crucial for converting plausible regional hypotheses into accurate fine-grained predictions, confirming that perception and verification are complementary. 

\paragraph{Effect of verification skills.}
We further ablate three verification skills: Linguistic Grounding (LG; OCR-based script/language cues), Scene Verification (SV; search-engine evidence), and Contextual Validation (CV; Map/POI evidence). As shown in Table~\ref{tab:ablation_results}, removing SV causes the largest drop, reducing Acc@500m from 16.87\% to 11.87\%, which indicates the importance of search-based scene grounding. Removing CV decreases Acc@500m to 13.58\% and Acc@750km to 84.31\%, showing that map-side context helps filter spatially inconsistent hypotheses. Removing LG leads to a milder but consistent decline, suggesting that OCR-derived script and language cues provide useful textual anchors. These results show that the three skills contribute complementary evidence for verification.

\paragraph{Effect of hypothesis budget.}
We additionally evaluate $K_{\max}=4$ on PAVED to examine the diversity--cost trade-off. Compared with the default $K_{\max}=3$ setting reported in Table~\ref{tab:tooluse}, $K_{\max}=4$ achieves 15.04\%, 25.06\%, 61.84\%, 70.25\%, and 85.53\% at 500\,m, 2\,km, 25\,km, 200\,km, and 750\,km, respectively. It improves performance at 25\,km and 200\,km but performs worse at the other thresholds, suggesting that additional hypotheses can improve regional candidate coverage while introducing distractors for fine-grained verification. We therefore retain $K_{\max}=3$ as the default setting in main experiments.

\vspace{-5pt}
\subsection{Quantitative Analysis (RQ3)}
  
\begin{figure}[t]
    \centering
    \includegraphics[width=\linewidth]{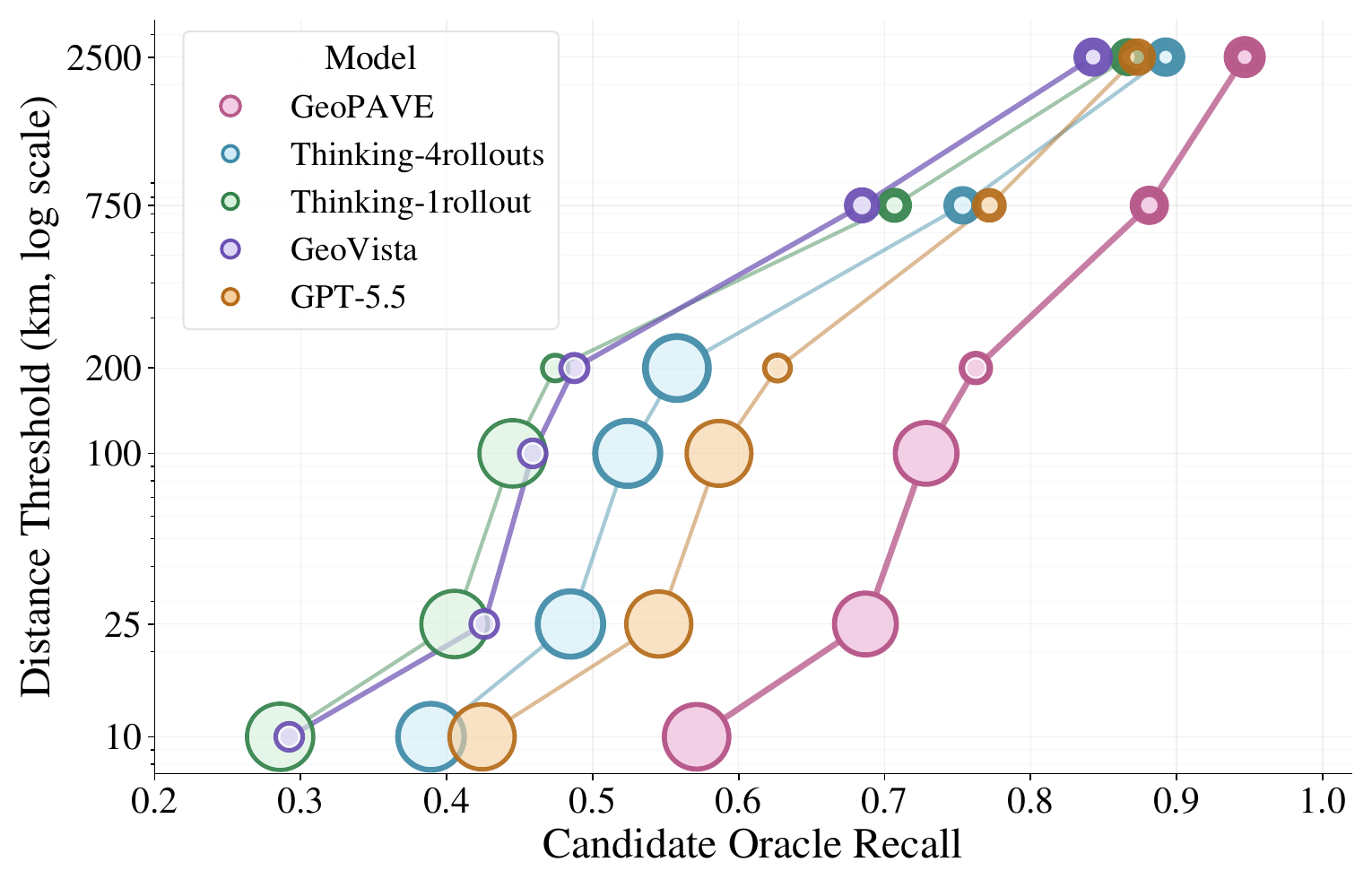}
    \caption{Diagnostic Analysis of Candidate Coverage and Evidence-Grounded Correction.}
    \label{fig:Diagnostic-analysis}
\end{figure}

\paragraph{Candidate Coverage and Correction.}
To examine candidate refinement and contextual consistency, we visualize tool-augmented models in Figure~\ref{fig:Diagnostic-analysis} with three diagnostic metrics: Candidate Oracle Recall (COR) for initial hypothesis coverage, Correction Recall (CR) for recovery from failed reasoning, and Log Correction Gain (LCG) for relative error reduction. Unlike Acc@$r$, these metrics expose the correction process behind final predictions. Figure~\ref{fig:Diagnostic-analysis} shows that baselines mainly improve at coarse thresholds, while GeoPAVE achieves higher COR and stronger fine-grained candidate coverage, especially within 10 km. These results suggest that GeoPAVE mitigates context drift by grounding hypothesis refinement in evidence-supported candidate regions.

\paragraph{Tool-Use Efficiency.} 
\begin{table}[t]
    \centering
    \small
    \setlength{\tabcolsep}{3pt}
    \begin{tabularx}{\columnwidth}{@{}l c >{\centering\arraybackslash}X c@{}}
        \toprule
        Method & Rollouts & Tools & Acc. (\%) \\
        \midrule
        GeoVista
            & 1
            & Web search
            & 1.55 / 39.22 \\

        \makecell[l]{Thinking\\w/ Map}
        & 4
        & Geocode, POI, Map
        & 9.09 / 51.03 \\

        GeoPAVE
            & 1$^\ast$
            & OCR, Web, Map, POI
            & \textbf{16.87 / 59.96} \\
        \bottomrule
    \end{tabularx}
    \caption{Comparison of inference structure, tool access, and
    geo-localization accuracy on PAVED. Accuracy is reported at
    500m/25km. $^\ast$GeoPAVE uses
    perception-verification process with $K_{\max}=3$ and
    $L_{\max}=3$.}
    \label{tab:tooluse}
\end{table} 
Table~\ref{tab:tooluse} summarizes the inference structure and external tools available to each method. GeoVista primarily relies on web search, while Thinking-with-Map uses four parallel rollouts augmented with geocoding, POI, and static-map tools. In contrast, GeoPAVE integrates OCR, web retrieval, and Map/POI evidence within a single perception-verification process, achieving 16.87\% and 59.96\% accuracy at 500m and 25km, respectively. These results suggest that structured multi-source verification helps mitigate perceptual shortcuts and prevent contextual inconsistency during long-horizon reasoning under visual ambiguity.

\vspace{-5pt}
\subsection{Case Studies}

\begin{figure}[!t]
    \centering
    \includegraphics[width=\linewidth]{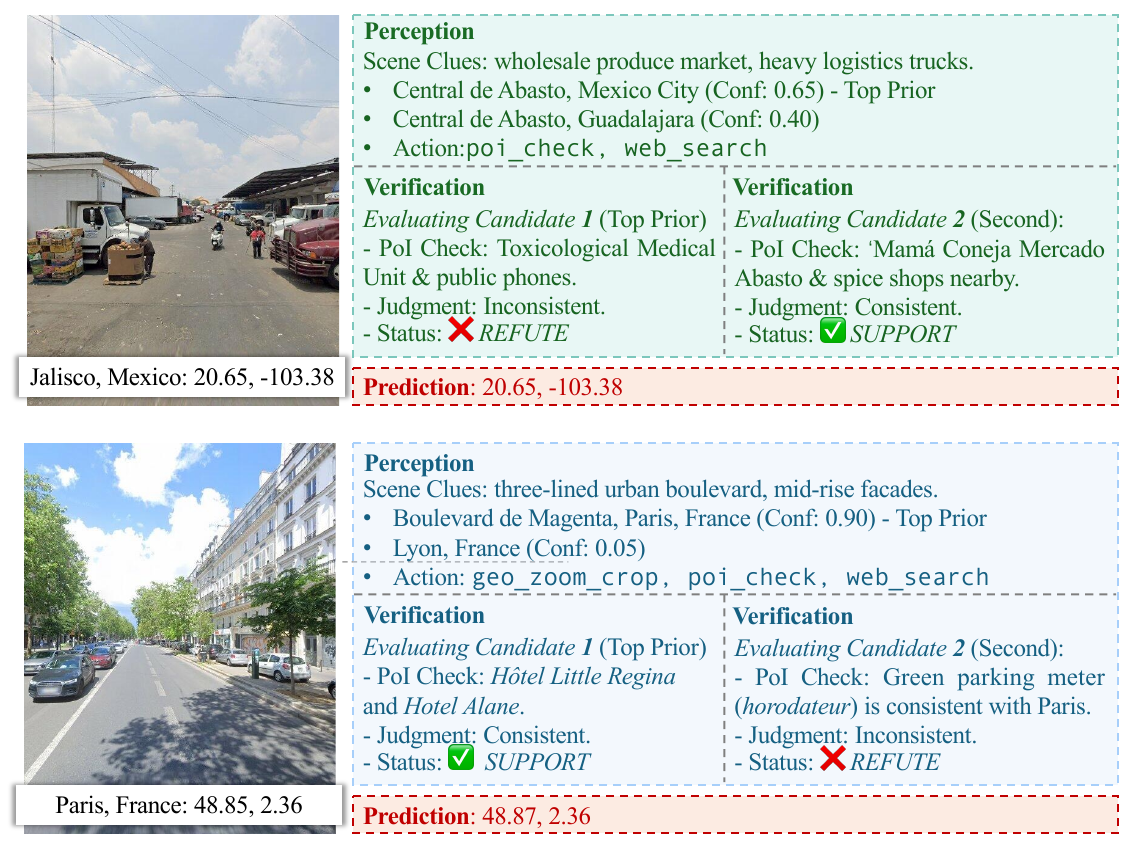}
    \caption{Case studies of evidence-grounded hypothesis
verification.}
    \label{fig:case}
\end{figure}

Figure~\ref{fig:case} shows two representative cases of the perception-verification process. In the first case, the perception stage recognizes the scene as a wholesale market with logistics trucks, and keeps both Mexico City and Guadalajara as plausible ``Central de Abasto'' candidates. Although the top prior points to Mexico City, the correct Guadalajara region remains in the candidate set, showing that the initial hypotheses provide useful coverage under visual ambiguity. The verification stage then checks nearby POI evidence. The Mexico City candidate is rejected because its local context is inconsistent with the observed market scene, while the Guadalajara candidate is supported by nearby market-related POIs, leading to the final correction.

The Paris case shows a complementary situation. The perception stage already proposes Boulevard de Magenta as the strongest hypothesis, and verification confirms it with nearby hotel and street-level evidence, while the Lyon alternative is ruled out due to inconsistent contextual cues. Together, these cases show that perception preserves plausible candidates and verification uses external evidence to correct or confirm them, reducing reliance on a single perceptual prior. Other detailed analysis information are provided in Appendix~\ref{sec:Qualitative} and \ref{sec:failure_analysis}.

\section{Conclusion and Future Works}

We reformulate open-world geo-localization as a \textbf{perceive-then-verify} reasoning problem and propose \textbf{GeoPAVE}, a bi-level agentic framework that combines perception-based hypothesis generation with evidence-grounded verification, reducing AI shortcuts and context inconsistency under ambiguous visual cues and demonstrating the effectiveness of GeoPAVE in both fine-grained geo-localization.

Future work will extend GeoPAVE and PAVED into a scalable data generation and training pipeline, using verified reasoning traces to train stronger geo-localization agents. We will further explore agentic reinforcement learning in interactive environments, where models can actively gather evidence and improve geographic reasoning through feedback.

\section{Acknowledgements}

This work is supported by the Guangdong Basic and Applied Basic Research Foundation (No. 2025A1515011994), the National Natural Science Foundation of China (No. 62402414), Guangdong Provincial Project 2025D03J0014, Guangzhou Municipal Science and Technology Project (No. 2023A03J0011), the Guangzhou Industrial Information and Intelligent Key Laboratory Project (No. 2024A03J0628), and Guangdong Provincial Key Lab of Integrated Communication, Sensing and Computation for Ubiquitous Internet of Things (No. 2023B1212010007).

\clearpage
\section*{Limitations}
While GeoPAVE demonstrates strong performance across multiple benchmarks, several limitations remain. First, the current verification module relies on a rule-based uncertainty-to-skill routing policy, which may not generalize optimally to edge cases where uncertainty signals are ambiguous or underspecified. 
Second, GeoPAVE's inference pipeline involves multiple sequential tool invocations, which introduces additional latency compared to single-pass reasoning approaches; future work could explore more efficient verification scheduling strategies. Third, although PAVED provides globally distributed coverage, its images are sourced from Foursquare check-ins and may over-represent commercially active urban areas, potentially limiting evaluation diversity in rural or less-documented regions. Finally, the trajectory curation process relies on a stronger reviewer model to filter low-quality traces, and the quality of PAVED annotations is thus partially dependent on the capabilities of this external model.

\section*{Ethical Consideration}

PAVED is derived from real-world check-in locations and street-view imagery, which raises privacy, licensing, and misuse concerns. We remove user identifiers, timestamps, and other check-in metadata that could reveal individual mobility patterns. When raw imagery cannot be redistributed under third-party licenses, we release only derived annotations, metadata, or retrieval scripts. Image geo-localization systems may be misused for surveillance, doxxing, or sensitive-location identification; therefore, GeoPAVE and PAVED are intended for benchmark research and should not be used to identify private individuals or infer sensitive personal locations.

AI-assisted technologies were used solely for language polishing and grammar correction to improve the clarity and readability of the manuscript. They were not used to generate the core ideas, experimental design, results, analyses, or references. The authors carefully reviewed all AI-assisted edits and take full responsibility for the integrity of the work, including the validity of the experimental results and the accuracy of all citations.

\bibliography{latex/custom}

\clearpage
\appendix

\twocolumn[{
    \renewcommand\twocolumn[1][]{#1}
    \begin{center}
      \textbf{\fontsize{15}{48}\selectfont Appendix}
    \end{center}
    \vspace{0.5cm}
}]

\label{sec:appendix}

\section{Experiment Results}
\label{sec:paved-rsults}
To provide a more fine-grained evaluation of GeoPAVE on PAVED, we further report Acc@10km and Acc@100km in addition to the main thresholds. These two intermediate thresholds measure city-level and regional-level localization accuracy, respectively, and help clarify whether the gains of GeoPAVE remain consistent beyond street-level prediction. As shown in Table~\ref{tab:paved-results}, GeoPAVE achieves 47.76\% at Acc@10km and 65.31\% at Acc@100km, outperforming all baselines at both thresholds. Compared with the strongest closed-source baseline, Gemini-3.1-Pro, GeoPAVE improves Acc@10km by 2.26 percentage points and Acc@100km by 4.61 percentage points. These results suggest that evidence-grounded verification not only improves fine-grained geo-localization, but also provides stable benefits for broader city- and region-level geographic reasoning.


\definecolor{apigreen}{RGB}{240, 249, 242}
\definecolor{geopavegreen}{RGB}{220, 242, 225}
\definecolor{apiblue}{RGB}{242, 247, 255}
\definecolor{geopaveblue}{RGB}{230, 240, 255}

\begin{table*}[t]
\centering
\small
\renewcommand{\arraystretch}{1.3}
\setlength{\tabcolsep}{4pt}

\begin{tabularx}{\textwidth}{l CCCCCCC}
\toprule
\multirow{2}{*}{\textbf{Model}} &
\multicolumn{7}{c}{\textbf{Performance on PAVED (\%)}} \\
\cmidrule(lr){2-8}
& 500m & 2km & 10km & 25km & 100km & 200km & 750km \\
\midrule

GeoCLIP
& 2.58 & 7.49 & 16.36 & 24.52 & 32.21 & 37.08 & 59.31 \\

GeoReasoner
& 0.99 & 5.23 & 21.35 & 31.42 & 34.07 & 36.57 & 50.71 \\

GAEA
& 1.86 & 7.25 & 24.60 & 34.39 & 38.19 & 41.64 & 61.69 \\

GeoVista
& 1.55 & 6.74 & 26.78 & 39.22 & 42.31 & 44.97 & 63.31 \\

GLOBE
& 0.67 & 2.85 & 10.34 & 14.86 & 18.23 & 20.80 & 38.99 \\

Thinking~with~Map
& 3.69 & 9.65 & 28.97 & 40.53 & 44.46 & 47.38 & 70.55 \\

Thinking~with~Map*4+verifier
& 9.09 & 17.89 & 38.46 & 51.03 & 55.29 & 58.07 & 78.30 \\

\midrule

Gemini-3.1-Pro
& 11.10 & 17.00 & 45.50 & 57.50 & 60.70 & 63.40 & 77.20 \\

GPT-5.5
& 5.31 & 13.16 & 30.06 & 39.42 & 43.55 & 47.28 & 67.51 \\

\rowcolor{apigreen}
GPT-4o-mini
& 2.77 & 10.62 & 26.39 & 34.90 & 38.11 & 40.77 & 52.93 \\

\rowcolor{apiblue}
Gemini-3.0-Flash
& 14.92 & 23.34 & 45.96 & 54.74 & 62.84 & 64.45 & 78.80 \\

\midrule

\rowcolor{geopavegreen}
GeoPAVE (GPT-4o-mini)
& 2.87 & 11.36 & 29.46 & 36.84 & 42.30 & 45.05 & 62.28 \\

\rowcolor{geopaveblue}
GeoPAVE (Gemini-3.0-Flash)
& \textbf{16.87} & \textbf{26.89} & \textbf{47.76}
& \textbf{59.96} & \textbf{65.31} & \textbf{69.35}
& \textbf{85.86} \\

\bottomrule
\end{tabularx}

\caption{Comparison of geolocation accuracy on PAVED under different
error thresholds (values in \%). Best results are shown in bold.
GeoPAVE is evaluated with GPT-4o-mini and Gemini-3.0-Flash as its
respective base models.}

\label{tab:paved-results}
\end{table*}

\section{Datasets}
\label{sec:appendix_datasets}

To ensure distributional diversity and rigorously evaluate the open-world geolocalization capabilities of agentic models, our evaluation spans three distinct datasets. These datasets were carefully selected and curated to assess models across legacy internet photos, region-specific urban environments, and human-centric reasoning scenarios.
\subsection{PAVED}
\label{sec:appendix_PAVED}

PAVED (Perception-and-Verification-Engine-Dataset) consists of 2,524 manually filtered street-view images collected from worldwide human check-ins, with hypothesis locations sourced from the Foursquare check-in corpus~\cite{foursquare}. Each image is anchored to a real point of interest visited by users and corresponds to accessible, semantically meaningful urban scenes, such as commercial streets, local neighborhoods, storefronts, road signs, and surrounding POIs. We retrieve high-resolution imagery through Google Maps and manually filter out non-localizable samples, including indoor scenes, close-up objects, visually uninformative views, and overly iconic global landmarks. As a result, PAVED focuses on realistic geo-localization cases where models must infer locations from rich but often ambiguous street-level semantic cues.

As shown in Table~\ref{tab:benchmark}, PAVED differs from existing geo-localization benchmarks in both data construction and reasoning supervision. While prior datasets often emphasize either global coverage, retrieval-style evaluation, or final-coordinate prediction, PAVED jointly provides global human-centric imagery, reasonable geo-localizability, high-resolution visual evidence, data diversity, and nuanced distance/address-level evaluation. More importantly, PAVED is paired with structured Chain-of-Thought annotations that decompose each sample into hypothesis generation and tool-augmented verification. These trajectories record how external evidence is retrieved, evaluated, and used to refine hypothesis locations, enabling PAVED to evaluate whether agentic models can actively verify hypotheses rather than rely only on parametric priors.

Unlike existing benchmarks that rely on randomly scraped coordinates or easily recognizable landmarks~\cite{recognition-globe}, PAVED is constructed from the Foursquare check-in corpus~\cite{foursquare}, ensuring that every image is anchored to genuine human activity and accessible, semantically meaningful places (e.g., commercial districts, local neighborhoods). All corresponding high-resolution imagery is retrieved via Google Maps to support reliable visual clue extraction and grounding. To maintain meaningful gep-localization difficulty, we conducted localizability filtering to remove non-localizable images (e.g., indoor rooms, close-up objects) and easily recognizable global landmarks. Most importantly, PAVED is accompanied by Structured Chain-of-Thought (Structured CoT) annotations that decouple the reasoning process into a hypothesis-generation state and a tool-augmented verification state. This unique structure provides a robust testbed to evaluate whether agentic models can actively retrieve external information to validate or refine their hypotheses, rather than relying solely on parametric priors.

\begin{figure}
    \centering
    \includegraphics[width=\linewidth]{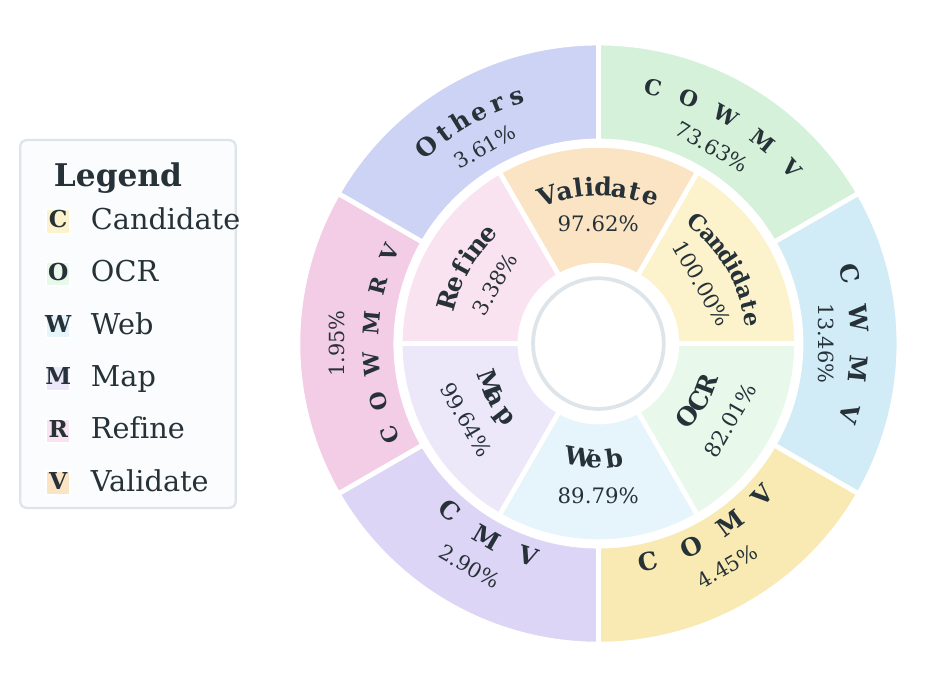}
    \caption{Tool Invocation Patterns in PAVED Verification Traces.}
    \label{fig:bubble}
\end{figure}

\begin{figure}[t]
    \centering
    \includegraphics[width=\linewidth]{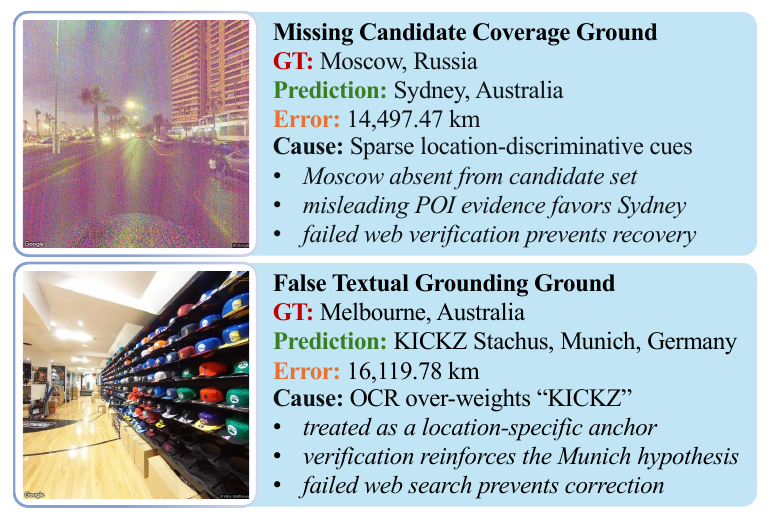}
    \caption{Representative GeoPAVE failures caused by missing candidate coverage and false textual grounding}
    \label{fig:failure_cases}
\end{figure}

\subsection{Details of Benchmarks}
\label{sec:appendix_datasets}
\paragraph{IM2GPS3K.} 
As a standard and widely recognized global benchmark~\cite{im2gps3k}, IM2GPS3K contains 2,997 geotagged images sourced from Flickr. This dataset serves to evaluate general, open-world geo-localization capabilities on legacy internet photos. Because these images are largely crowd-sourced and vary significantly in quality, visual composition, and geographic context, they pose a traditional yet robust challenge for testing whether models can extract meaningful location cues from unstructured and often noisy visual data.

\paragraph{MAPBench-V2.}
To test fine-grained, region-specific reasoning and robustness, we utilize the 2.4K-image test split of MAPBench-V2~\cite{amap}. This benchmark is heavily China-centered, featuring a diverse array of urban points of interest (POIs) and street-view scenes across multiple cities. Evaluating on MAPBench-V2 allows us to assess a model's ability to interpret specific regional infrastructure, localized textual cues (e.g., Chinese storefront signage), and distinct architectural styles, thereby testing the limits of its geographic knowledge and fine-scale urban geolocalization.

\subsection{Baselines}
\label{sec:appendix_Baselines}

\begin{figure*}[ht]
    \centering
    \includegraphics[width=1\textwidth]{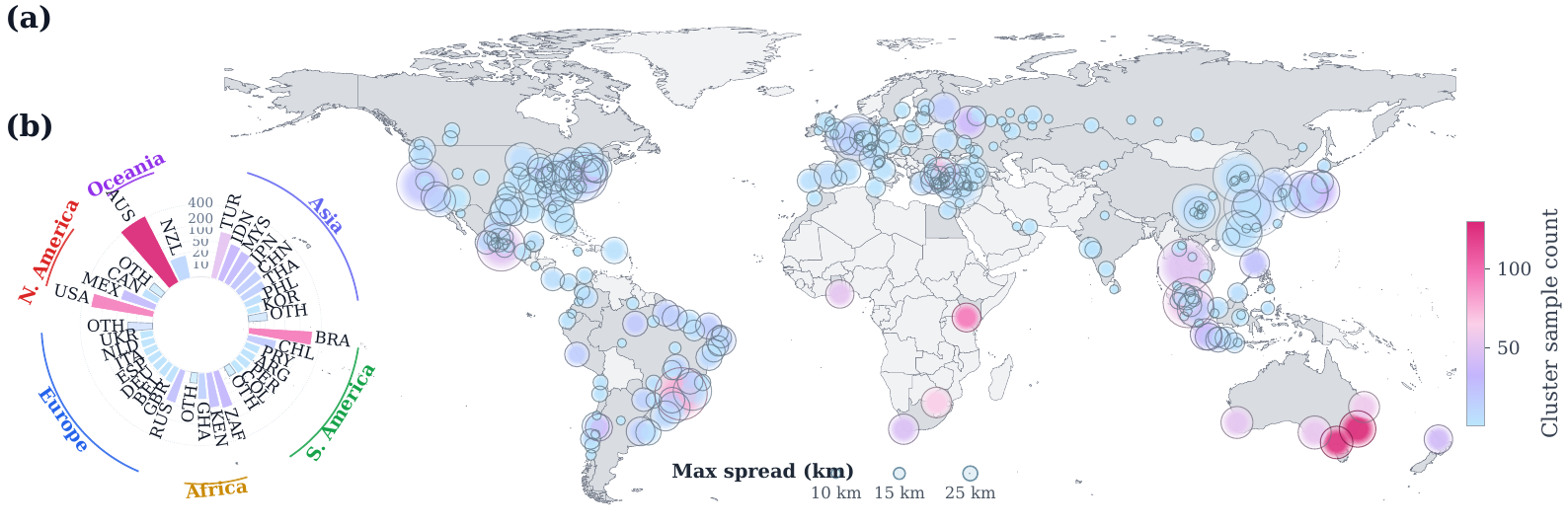}
    \caption{Geographical Distribution of PAVED. Our dataset provides high-resolution, globally annotated imagery sourced from real human check-ins, designed to evaluate the general perception and verification abilities of agentic models.}
    \label{fig:worldmap}
\end{figure*}
\subsubsection*{Open-source Models}
\begin{itemize}[leftmargin=*]
    \item \textbf{GeoCLIP~\cite{vivanco2023geoclip}:} A CLIP-inspired image-to-GPS retrieval approach that models the Earth as a continuous space. It employs positional encoding with random Fourier features and constructs a hierarchical representation to effectively align image embeddings with corresponding GPS locations for worldwide geolocalization.
    
    \item \textbf{GeoReasoner~\cite{li2024georeasoner}:} A novel framework leveraging large vision-language models (LVLMs) augmented with human inference knowledge derived from real geo-localization games. To ensure data quality, it first curates a dataset of highly locatable street views, followed by specialized fine-tuning through dedicated reasoning and location-tuning stages to enhance geographical deduction.
    
    \item \textbf{GAEA~\cite{campos2025gaea}:} A conversational visual-language model explicitly designed for geolocation awareness. It is trained on GAEA-1.4M, a synthesized large-scale dataset built upon OpenStreetMap attributes, enabling nuanced location-specific reasoning and dialogue beyond pure coordinate prediction.
    
    \item \textbf{GeoVista~\cite{wang2025geovista}:} A web-augmented agentic model designed for geolocalization that integrates explicit tool invocations, such as image zooming and web searching, directly into the visual reasoning loop. It is optimized through supervised fine-tuning followed by reinforcement learning.

\item \textbf{smileGeo~\cite{han2025swarm}:} A collaborative swarm intelligence framework that leverages multiple Internet-enabled LVLM agents to associate images with precise geographic locations. It facilitates inter-agent communication to synthesize models' inherent knowledge with retrieved web information, employing a dynamic learning strategy to optimize interactions and reduce redundant communication.

    \item \textbf{Thinking~with~Map~\cite{amap}:} A map-augmented reasoning agent that utilizes reinforcement learning to gather geographic evidence. To comprehensively evaluate its capacity, we benchmark it under two distinct inference settings:
    \begin{itemize}[leftmargin=*]
        \item \textbf{1 rollout:} The base setting where the RL-finetuned agent is directly utilized to perform a single-pass inference for coordinate prediction.
        \item \textbf{4 rollouts:} A parallel test-time scaling framework where four agents generate independent hypothesis locations. In our implementation, considering the trade-off between reasoning performance and computational overhead, we adapted the original pipeline by deploying Gemini 3.0 Flash as the strict verifier (in place of the original Qwen3-VL-235B-A22B) to synthesize the gathered evidence and output the most geographically plausible final coordinate.
    \end{itemize}
\end{itemize}

\subsubsection*{Closed-source Models}
\begin{itemize}[leftmargin=*]
    \item \textbf{Gemini 3.1 Pro:} State-of-the-art native multimodal foundation model developed by Google. Thanks to its massive context windows and advanced high-resolution visual processing capabilities, Gemini 3.1 Pro excels at extracting fine-grained geographic clues (e.g., street signs, architectural styles, and vegetation), serving as a heavy-weight baseline offering deep, multi-step spatial reasoning and extensive zero-shot world knowledge.
    
    \item \textbf{GPT-5.5:} The latest proprietary large multimodal model introduced by OpenAI, representing the current frontier of closed-source visual reasoning. It demonstrates unprecedented zero-shot generic visual comprehension and cross-lingual text-in-image reading capabilities. For geolocalization, it leverages powerful chain-of-thought (CoT) reasoning to synthesize diverse environmental clues and accurately deduce global coordinates even in unconstrained or ambiguous street views.
\end{itemize}

\section{Evaluation Metric Details}
\label{sec:appendix_Metrics}

This section supplements the evaluation metrics defined in Section~5.1. 
We provide implementation details for coordinate parsing, distance computation, candidate-based diagnostics, and aggregation.

\paragraph{Coordinate accuracy.}
For each image $I_i$, a method outputs a final coordinate $\hat{p}_i=(\widehat{\mathrm{lat}}_i,\widehat{\mathrm{lon}}_i)$. 
If a response contains multiple coordinates, we use the coordinate explicitly marked as the final answer. 
For GeoPAVE, this corresponds to the final prediction after verification. 
For baselines, we parse the final latitude--longitude pair using a deterministic coordinate parser.

We compute the geographic error using the haversine distance with Earth radius $R=6371.0088$ km. 
Distance-threshold accuracy is defined as:
\[
\mathrm{Acc}@r =
\frac{100}{N}
\sum_{i=1}^{N}
\mathbf{1}
\left[
d(\hat{p}_i,p_i^\ast)\le r
\right].
\]
We report $\mathrm{Acc}@r$, where a prediction is considered correct if its haversine distance to the ground-truth coordinate is within radius $r$. We evaluate $r \in \{0.5, 2, 10, 25, 100, 200, 750\}$ km, and present representative thresholds in the main tables due to space constraints. Predictions with missing, invalid, or unparsable coordinates are counted as incorrect for all thresholds.

\paragraph{Candidate-based diagnostics.}
For GeoPAVE, the initial candidate set $H_i^{(0)}$ is the set of hypotheses generated by the perception module before verification. 
The initial prediction $\tilde{p}_i$ is the top-ranked candidate before verification, and the final prediction $\hat{p}_i$ is the coordinate selected after evidence-grounded refinement. 
For methods that output only one coordinate, the initial candidate set is treated as a singleton.

We use the same definitions of Candidate Oracle Recall (COR), Correction Recall (CR), and Log Correction Gain (LCG) as in Section~5.1. 
For CR, we set $\tau=r$ unless otherwise specified, meaning that $\mathrm{CR}@r$ measures the fraction of samples that are initially wrong at threshold $r$ but become correct after verification. 
If no sample fails initially at threshold $r$, CR is reported as N/A. 
For LCG, invalid predictions are assigned a clipped maximum error of $20{,}015$ km, approximately half of Earth's circumference.


\section{Implementation Details}

\subsection{Perception}
\label{sec:appendix_perception}

The perception stage first produces a coarse fallback and three candidate locations, each with confidence, uncertainty tags, and recommended verification actions. We expose two image-level tools, Zoom-Crop and OCR, whose definitions are shown in Figure~\ref{fig:tool_zoom_crop} and Figure~\ref{fig:tool_ocr}. The controller maps uncertainty tags to tool actions through the routing policy in Figure~\ref{fig:alg_perception_routing}, and the OCR prompt is provided in Figure~\ref{fig:prompt_ocr_module}.

\begin{figure*}[t]
\begin{tcolorbox}[geotool,width=\textwidth,title=Tool: \texttt{Zoom-Crop}]
\textbf{Description:} Crops a complete local region for close inspection when a visible sign, storefront, plate, address panel, or landmark detail is present but not readable at full-image scale.

\textbf{Arguments:}
\begin{itemize}
    \item \texttt{bbox\_2d} (array of numbers): Bounding box in normalized image coordinates. \textit{(Required)}
    \item \texttt{label} (string): Short label for the target region. \textit{(Optional)}
\end{itemize}
\end{tcolorbox}
\caption{Perception-stage Zoom-Crop tool definition.}
\label{fig:tool_zoom_crop}
\end{figure*}

\begin{figure*}[t]
\begin{tcolorbox}[geotool,width=\textwidth,title=Tool: \texttt{OCR}]
\textbf{Description:} Reads legible text from the original image or a returned crop, and converts it into compact place-like and auxiliary text evidence.

\textbf{Arguments:}
\begin{itemize}
    \item \texttt{image\_id} (string): The original image or the crop identifier returned by \texttt{Zoom-Crop}. \textit{(Required)}
\end{itemize}
\end{tcolorbox}
\caption{Perception-stage OCR tool definition.}
\label{fig:tool_ocr}
\end{figure*}

\begin{figure*}[t]
\begin{tcolorbox}[geotool,width=\textwidth,title=\texttt{Algorithm: Perception Tool Routing},before upper={\small\sloppy\raggedright}]
\begin{algorithmic}[1]
\Require Image $I$, top candidate $c_1$, uncertainty tags $T$, recommended actions $A$, need-zoom flag $z$, zoom targets $Z$.
\State $T_{\mathrm{text}} \gets \{\texttt{TEXT\_UNCLEAR}, \texttt{SYMBOLS\_UNCLEAR}\}$.
\State $T_{\mathrm{text}} \gets T_{\mathrm{text}} \cup \{\texttt{NEED\_ZOOM}\}$.
\State $Z_{\mathrm{text}} \gets \{\texttt{sign\_text}, \texttt{license\_plate}, \texttt{storefront}\}$.
\State $T_{\mathrm{map}} \gets \{\texttt{CONTEXT\_MISSING}, \texttt{TOPOGRAPHY\_UNCLEAR}, \texttt{LANDMARK\_ID\_UNCLEAR}\}$.
\State $T_{\mathrm{map}} \gets T_{\mathrm{map}} \cup \{\texttt{WATERBODY\_DOMINANT}, \texttt{CONFLICTING\_CUES}\}$.
\State $r_{\mathrm{text}} \gets z$ or $T \cap T_{\mathrm{text}} \neq \emptyset$ or $Z \cap Z_{\mathrm{text}} \neq \emptyset$.
\State $r_{\mathrm{text}} \gets r_{\mathrm{text}}$ or \texttt{OCR} $\in A$ or \texttt{ZOOM} $\in A$.
\If{no visible text, sign, storefront, plate, or landmark detail exists}
    \State Skip image tools.
\ElsIf{visible text is already readable}
    \State Store it as OCR memory.
\ElsIf{$r_{\mathrm{text}}$}
    \State Crop the complete text-bearing object if a close-up is needed.
    \State Run OCR on the crop or original image.
\EndIf
\If{OCR memory contains a place, brand, road, or venue, or visual/text cues create an unresolved location conflict}
      \State Record a short reasoning Web query or POI query for Verification.
  \EndIf
\If{$T \cap T_{\mathrm{map}} \neq \emptyset$}
    \State Record a Map Context request for Verification.
\EndIf
\end{algorithmic}
\end{tcolorbox}
\caption{Perception-stage tool routing algorithm.}
\label{fig:alg_perception_routing}
\end{figure*}

\begin{figure*}[htbp]
\begin{tcblisting}{geoprompt,width=\textwidth,title=\texttt{Prompt: OCR Module},listing only}
You are an OCR module.
Read ALL legible text from the image.

Return ONLY a valid JSON object with this schema (no markdown, no extra text):
{
  "place_like_texts": ["...", "..."],
  "other_texts": ["...", "..."]
}

Rules:
- place_like_texts: ONLY snippets that look like names of places/venues/brands/organizations.
- other_texts: any other readable text (optional).
- Prefer short readable snippets over empty output when at least 3 consecutive visible characters can be read.
- If a snippet may be useful for geolocation but you are unsure whether it is a place/brand/organization, include it in other_texts instead of dropping it.
- If nothing readable, return empty lists.
- Do NOT invent text you cannot see.
\end{tcblisting}
\caption{OCR module prompt used in the perception stage.}
\label{fig:prompt_ocr_module}
\end{figure*}

\subsection{Verification}
\label{sec:appendix_verification}

The verification stage grounds each perception candidate with OCR, Web, and Map evidence. We define the verification tools in Figure~\ref{fig:tool_text_evidence}, Figure~\ref{fig:tool_web_evidence}, and Figure~\ref{fig:tool_map_context}. Map evidence is exposed through three compact sub-tools rather than raw map-service outputs. Tool invocation and refinement follow Figure~\ref{fig:alg_verification_routing}. All tool outputs are serialized into the compact interface in Figure~\ref{fig:interface_tool_evidence_serialization} before being sent to the verification refiner prompt in Figure~\ref{fig:prompt_verification_refiner}.

\begin{figure*}[t]
\begin{tcolorbox}[geotool,width=\textwidth,title=Tool: \texttt{Text Evidence}]
\textbf{Description:} Reuses OCR memory as image-grounded anchors for candidate validation and address refinement.

\textbf{Arguments:}
\begin{itemize}
    \item \texttt{ocr\_texts} (array of strings): Place-like and auxiliary snippets extracted from the image. \textit{(Required)}
\end{itemize}
\end{tcolorbox}
\caption{Verification-stage text evidence tool definition.}
\label{fig:tool_text_evidence}
\end{figure*}

\begin{figure*}[t]
\begin{tcolorbox}[geotool,width=\textwidth,title=Tool: \texttt{Web Evidence}]
\textbf{Description:} Executes perception-authored reasoning, OCR, and POI queries, then returns compact search evidence for each candidate.

\textbf{Arguments:}
\begin{itemize}
    \item \texttt{queries} (array of strings): Web or POI-oriented queries authored during perception. \textit{(Required)}
\end{itemize}
\end{tcolorbox}
\caption{Verification-stage web evidence tool definition.}
\label{fig:tool_web_evidence}
\end{figure*}

\begin{figure*}[t]
\begin{tcolorbox}[geotool,width=\textwidth,title=Tool: \texttt{Map Context}]
\textbf{Description:} Provides compact map-side evidence for a candidate location, including local POI context, terrain plausibility, and spatial feature consistency. The tool returns structured support/refute signals rather than raw map-service outputs.

\textbf{Arguments:}
\begin{itemize}
    \item \texttt{coordinate} (pair of floats): Candidate latitude and longitude. \textit{(Required)}
    \item \texttt{radius\_m} (integer): Local validation radius; default is 500m. \textit{(Optional)}
    \item \texttt{keyword} (string): Optional POI name or category from perception queries. \textit{(Optional)}
    \item \texttt{expected\_terrain} (string): Predicted terrain tag, e.g., coastal, flat, hilly, or alpine. \textit{(Optional)}
    \item \texttt{expected\_features} (array of strings): Visual features to verify, such as coastline, river, forest, peak, bridge, station, or park. \textit{(Optional)}
\end{itemize}

\textbf{Core Capabilities:}
\begin{itemize}
    \item \textbf{Neighborhood POI Probe:} counts nearby POIs and returns nearest POI names, types, and distances within the local radius.
    \item \textbf{Topographic Consistency Check:} compares external elevation evidence with the terrain/elevation tag predicted from the image.
    \item \textbf{Spatial Feature Cross-Check:} checks local map-feature counts and ring-style spatial evidence to detect contradictions with the visual scene.
\end{itemize}
\end{tcolorbox}
\caption{Verification-stage map context tool definition.}
\label{fig:tool_map_context}
\end{figure*}

\begin{figure*}[t]
\begin{tcolorbox}[geotool,width=\textwidth,title=\texttt{Algorithm: Verification Tool Routing},before upper={\small\sloppy\raggedright}]
\begin{algorithmic}[1]
\Require Initial hypotheses $Z^{(0)}=\{h_i^{(0)}\}_{i=1}^{K}$, perception output $P$, OCR memory $O$, tool availability $G$, budgets $B$.
\Ensure Evidence pool $\mathcal{E}$, refined hypotheses $Z^{(1)}$, and final prediction $\hat h$.
\Statex \textbf{Part I: Skill Invocation}
\State Initialize evidence pool $\mathcal{E}\gets\emptyset$ and query bank $Q\gets\emptyset$.
\State Append to $Q$: $P.\texttt{reasoning\_web\_queries}$, $P.\texttt{ocr\_web\_queries}$, $P.\texttt{poi\_search\_queries}$, and candidate-level Web/POI queries from $Z^{(0)}$.
\State Deduplicate $Q$, reject generic or malformed queries, and retain at most $B_Q$ admitted queries.
\State Let $U_1$ and $c_1^{(0)}$ denote the top-hypothesis uncertainty tags and confidence.
\If{Web Evidence is available and $Q$ contains admitted Web queries}
    \If{the Web circuit is open or the text-only path has no OCR anchor}
        \State Skip Web Evidence and record the skip reason.
    \Else
        \State Invoke Web Evidence on the top budgeted Web queries and add compact snippets/entities to $\mathcal{E}$.
    \EndIf
\EndIf
\State Select up to $B_V$ validation hypotheses from $Z^{(0)}$ by confidence, semantic specificity, and uncertainty; include low-confidence or conflict-marked candidates.
\For{each selected hypothesis $h_i^{(0)}$}
    \State Add Text Evidence from OCR memory $O$ to $\mathcal{E}_i$.
    \State Invoke Map Context for $h_i^{(0)}$ when map evidence is available or $U_i$ contains context, topography, landmark, waterbody, or conflicting-cue uncertainty.
    \State Add Neighborhood POI Probe within 500m, matched POI detail when an anchor matches, and topographic consistency evidence to $\mathcal{E}_i$.
    \State Serialize $h_i^{(0)}$, image summary, query bank, OCR memory, Web Evidence, and Map Context as $\mathrm{CONTEXT\_JSON}_i$.
\EndFor
\Statex \textbf{Part II: Refinement}
\If{no low-confidence, conflict, query, or external-evidence signal is present}
    \State Skip refinement and keep the current evidence-gated ranking.
\EndIf
\For{each serialized context $\mathrm{CONTEXT\_JSON}_i$}
    \State $j_i\gets\mathrm{VerificationRefiner}(\mathrm{CONTEXT\_JSON}_i)$.
    \State Parse $j_i$ as \{\texttt{confidence}, \texttt{consistent}, \texttt{reason\_short}, \texttt{refined\_address}, \texttt{conflicting\_cues}, \texttt{new\_candidates}\}.
    \If{$j_i.\texttt{conflicting\_cues}\neq\emptyset$}
        \State Cap $j_i.\texttt{confidence}<0.90$.
    \EndIf
    \State Add a self-consistency evidence item to $\mathcal{E}$ with decision \textsc{Support}, \textsc{Refute}, or \textsc{NoData}.
    \If{$j_i.\texttt{consistent}$ and $j_i.\texttt{confidence}\ge0.90$}
        \State Accept $h_i^{(0)}$ as validated; apply \texttt{refined\_address} only as an address refinement for the same hypothesis.
        \State \Return $Z^{(1)}$ and $\hat h=h_i^{(1)}$.
    \Else
        \State Keep $h_i^{(0)}$ unresolved and collect at most $B_S$ admissible proposals from \texttt{refined\_address} and \texttt{new\_candidates}.
    \EndIf
\EndFor
\For{each collected proposal $q_s$}
    \State Land $q_s$ as a new candidate, then repeat Map Context serialization and VerificationRefiner scoring.
    \If{the spawned candidate reaches confidence $\ge0.90$}
        \State Insert it into $Z^{(1)}$ and return it as $\hat h$.
    \EndIf
\EndFor
\State Select $\hat h=\arg\max_{h_i\in Z^{(1)}}c_i^{(1)}$ among evidence-gated candidates; if none is validated, keep the best unresolved score below $0.90$ and expose remaining conflicts.
\end{algorithmic}
\end{tcolorbox}
\caption{Verification-stage tool routing and refinement algorithm.}
\label{fig:alg_verification_routing}
\end{figure*}

\begin{figure*}[t]
\begin{tcblisting}{geoprompt,width=\textwidth,title=\texttt{Interface: Tool-Evidence Serialization},listing only}
Tool outputs are serialized as CONTEXT_JSON:
{
  "task": "Judge whether one candidate is self-consistent with image reasoning and tool evidence. Return JSON only.",
  "candidate": {...},
  "image_summary": {
    "text_presence": "...",
    "scene_clues": ["..."],
    "expected_elevation_tag": "...",
    "top_level_conflicting_cues": ["..."]
  },
  "query_bank": [{"q": "...", "src": "...", "web": true, "poi": false}],
  "ocr_texts": ["..."],
  "web_search": [{"query": "...", "title": "...", "url": "...", "snippet": "..."}],
  "web_entities": ["..."],
  "nearby": {"count": 0, "radius_m": 500, "pois": [...]},
  "poi_details": [{...}],
  "elevation_m": 0.0
}
\end{tcblisting}
\caption{Tool-evidence serialization interface passed to the verification refiner.}
\label{fig:interface_tool_evidence_serialization}
\end{figure*}

\begin{figure*}[t]
\begin{tcblisting}{geoprompt,width=\textwidth,title=\texttt{Prompt: Verification Refiner},listing only,
  listing options={
    breaklines=true,
    breakatwhitespace=false, % 允许在非空格处折行，防止 JSON 字符串被强行拉伸
    basicstyle=\ttfamily\small, % 强制指定使用等宽字体和小号字
    columns=flexible % 保持字符紧凑度
  }
}
Return ONLY a valid JSON object with this exact shape:
{"confidence":0.0,"consistent":false,"reason_short":"<=160 chars","refined_address":"","conflicting_cues":[],"new_candidates":[]}
Rules: confidence is this perception candidate's consistency after merging image reasoning/memory with new external evidence. If evidence supports refining the address for the SAME perception candidate, set refined_address. If confidence stays <0.90, propose a next candidate only when the evidence clearly supports switching; otherwise leave new_candidates empty and the controller will advance to the next perception candidate. If conflicting cues remain, confidence MUST be < 0.90. Use only provided evidence; do not invent street names or POIs. Do not force new candidates to be near the fallback; judge semantic consistency from the evidence.

CONTEXT_JSON:
{CONTEXT_JSON}
\end{tcblisting}
\caption{Verification refiner prompt.}
\label{fig:prompt_verification_refiner}
\end{figure*}

\subsection{Qualitative Results}
\label{sec:Qualitative}

\begin{figure*}[htbp]
    \centering
    \begin{minipage}[t]{0.34\textwidth}
        \vspace{0pt} 
        \imageheader{Johannesburg, South Africa}
        \begin{tcolorbox}[imageframebox]
            \includegraphics[width=\linewidth]{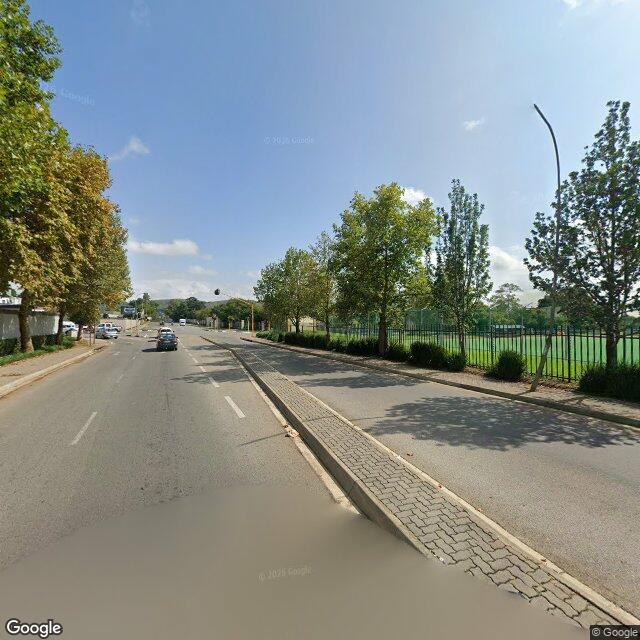}
        \end{tcolorbox}
        
        \vspace{2.2mm} 
        
        \modelheader{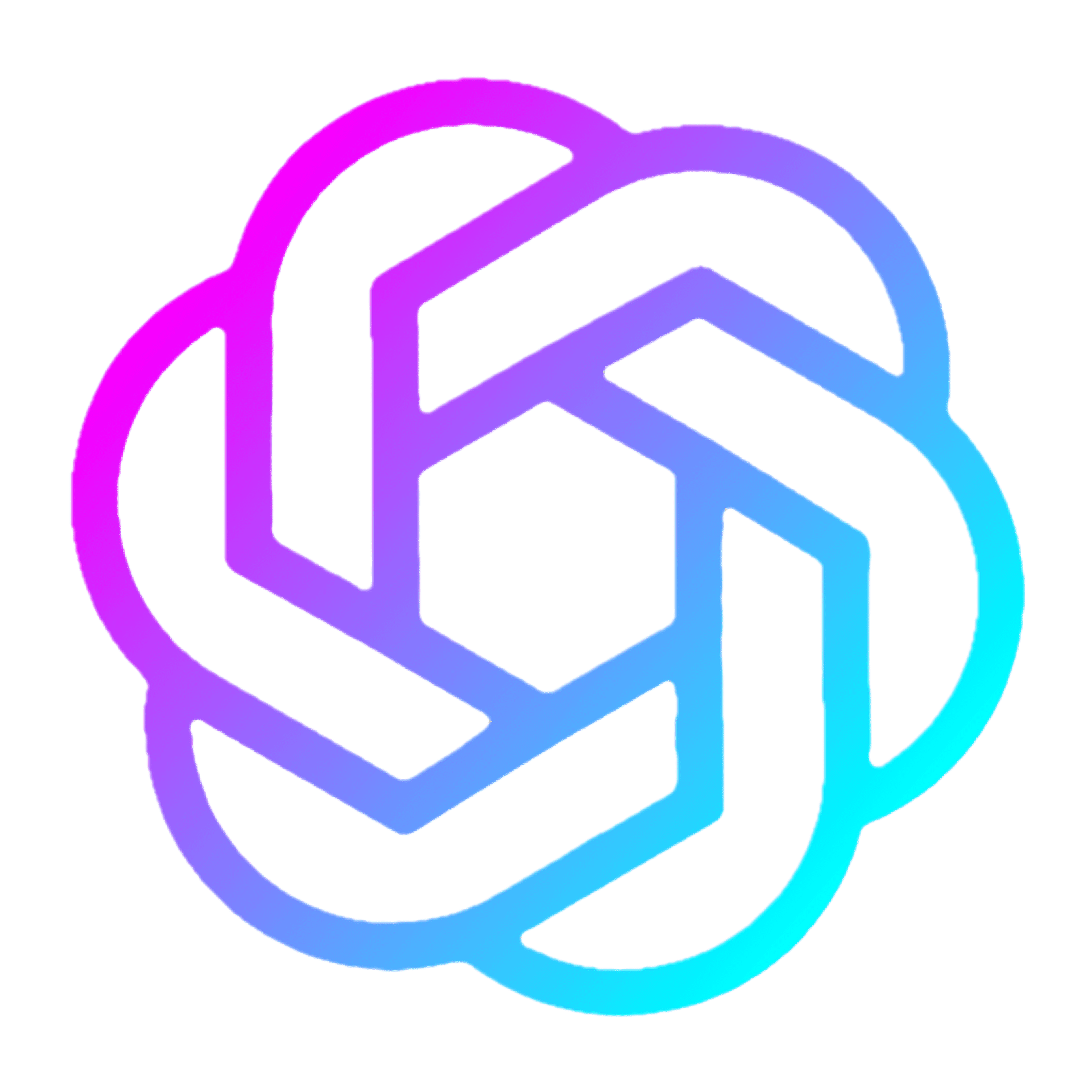}{GPT-5.5}{\textcolor{graytext}{Luxembourg, Luxembourg}}
        \begin{tcolorbox}[modelboxgpt]
            \traceitem{Visual premise:}{Calls the scene ``Luxembourg City'' with \textcolor{errortext}{right-hand traffic}, \textcolor{errortext}{European road furniture}, \textcolor{errortext}{Luxembourg-style curbs/paving}, and a \textcolor{highlighttext}{fenced sports field}.}
            \traceitem{Place prior:}{Maps the tree-lined sports-ground scene to \textcolor{errortext}{Limpertsberg/Belair} near \textcolor{errortext}{Avenue de la Fa\"iencerie} and \textcolor{errortext}{Val Sainte-Croix}.}
            \traceitem{Failure mode:}{The wrong traffic-direction cue compounds into a European location prior.}
            \traceitem{Final:}{\textcolor{errortext}{Luxembourg, Luxembourg}; error 8706.31 km \myemoji{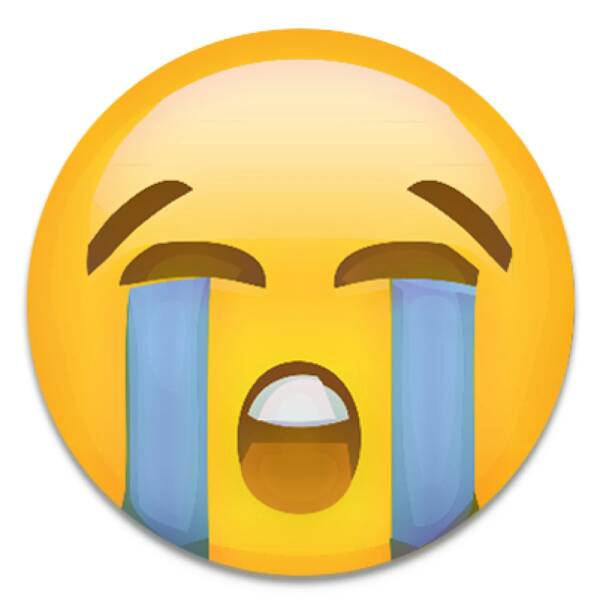}.}
        \end{tcolorbox}
    \end{minipage}%
    \hfill
    \begin{minipage}[t]{0.64\textwidth}
        \vspace{0pt} 
        
        \modelheader{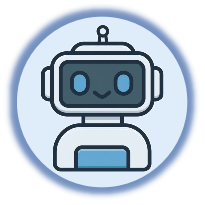}{GeoPAVE}{Johannesburg, South Africa}
        \begin{tcolorbox}[modelboxtop]
            \traceitem{Perception:}{Extracts \textcolor{highlighttext}{left-hand driving}, \textcolor{highlighttext}{yellow road edge lines}, \textcolor{highlighttext}{interlocking brick median paving}, \textcolor{highlighttext}{green sports-field fencing}, and \textcolor{highlighttext}{hilly Highveld-like terrain}; no decisive text is readable.}
            \traceitem{Routing:}{Marks \texttt{TEXT\_UNCLEAR} and \texttt{GENERIC\_SCENE}, routing the case from direct prediction to Web, POI, and map-context verification.}
            \traceitem{Hypotheses:}{Maintains Johannesburg (0.60), Pretoria (0.30), and Durban (0.10), preserving candidate diversity under ambiguity.}
            \traceitem{Tools:}{Issues the cue-aligned Web queries and candidate-specific POI/map checks described in the method.}
            \traceitem{Evidence pool:}{Serializes Johannesburg evidence at \textcolor{highlighttext}{Sandhurst/Sandton} and Pretoria evidence at \textcolor{highlighttext}{Pretoria Central} together with nearby POI and terrain context.}
            \traceitem{Refiner:}{Re-scores the candidate set against the evidence pool; both landed candidates are downgraded to 0.30 consistency.}
            \traceitem{Decision:}{Applies \textsc{Refute} to the precise candidates and keeps the broader Johannesburg fallback because dense CBD contexts conflict with the hilly sports-field scene.}
            \traceitem{Final:}{Falls back to the broader \textcolor{highlighttext}{Johannesburg} hypothesis; error 8.24 km \myemoji{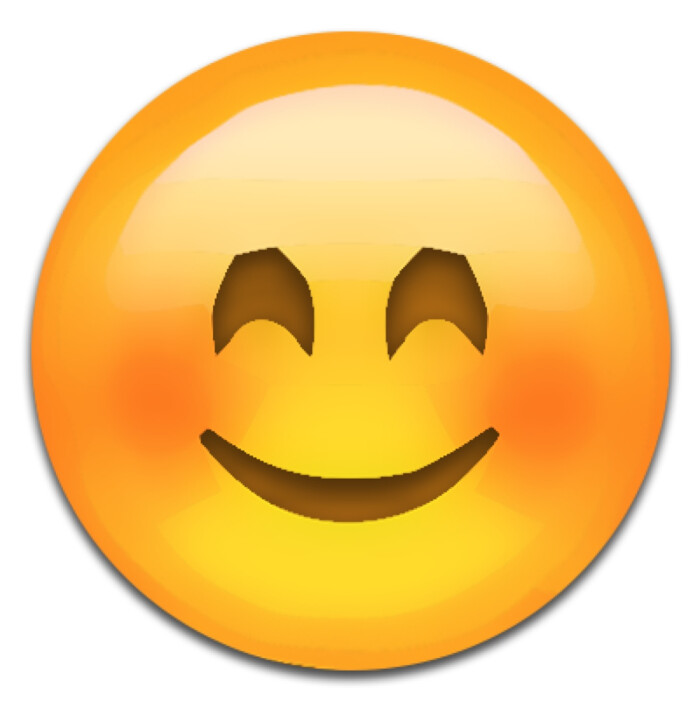}.}
        \end{tcolorbox}
        
        \vspace{2.2mm}
        
        \modelheader{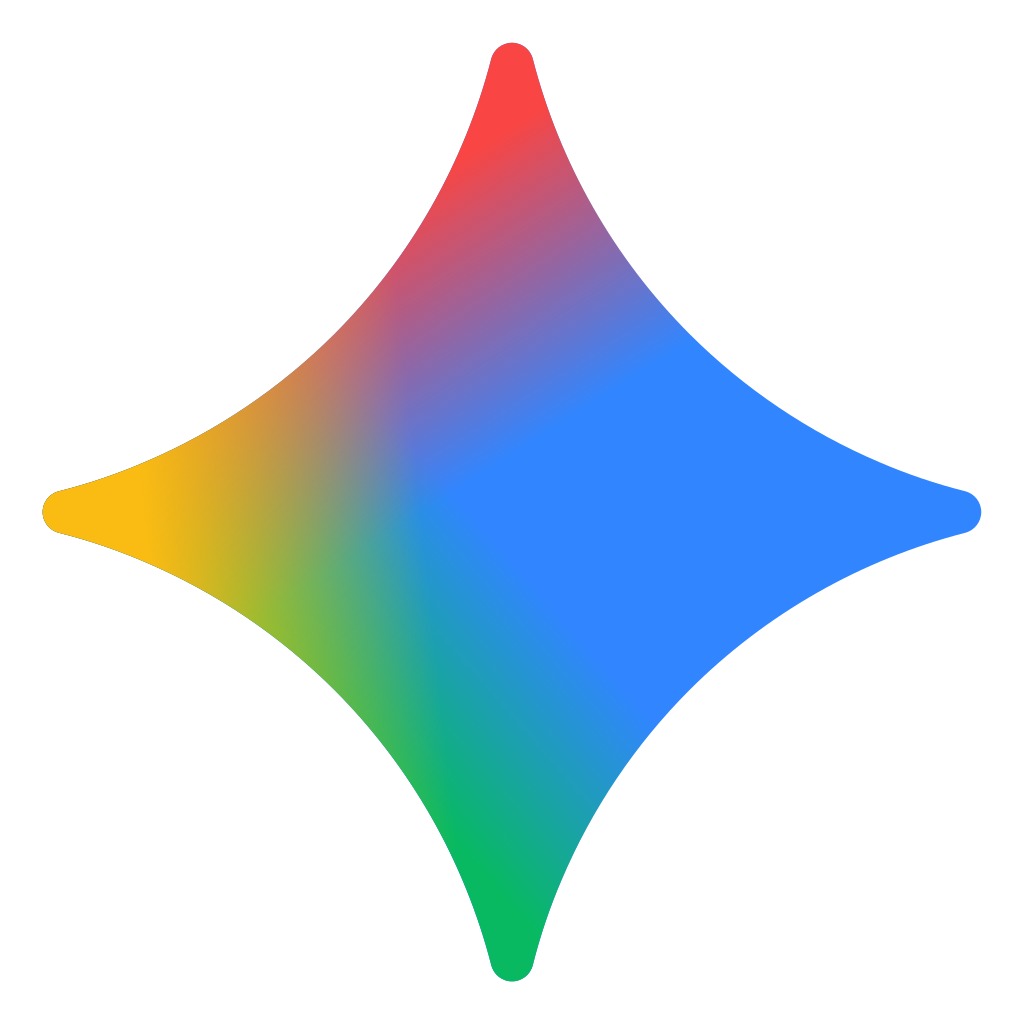}{Gemini-3.1-Pro}{\textcolor{graytext}{Pretoria}, South Africa}
        \begin{tcolorbox}[modelboxbottom]
            \traceitem{Perception:}{Correctly identifies \textcolor{highlighttext}{left-side driving}, \textcolor{highlighttext}{interlocking zig-zag bricks}, \textcolor{highlighttext}{green palisade-style fencing}, \textcolor{highlighttext}{curved streetlights}, and \textcolor{highlighttext}{rolling Highveld terrain}.}
            \traceitem{Scene layout:}{Uses tall curved streetlights and the general urban/suburban road layout as \textcolor{highlighttext}{South African context}.}
            \traceitem{Ambiguity:}{Notes that the terrain suggests either Pretoria or Johannesburg, but \textcolor{errortext}{does not verify the city-level choice.}}
            \traceitem{Final:}{Stops after single-pass reasoning and selects \textcolor{errortext}{Pretoria}; error 50.24 km \myemoji{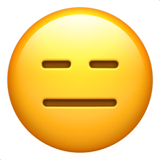}.}
        \end{tcolorbox}
    \end{minipage}
    
    \caption{Reasoning comparison of three different models (GeoPAVE, Gemini-3.1-Pro, and GPT-5.5) on the same input image. Reliable visual cues identified by the models are marked in \textcolor{highlighttext}{blue}, while severe visual hallucinations are marked in \textcolor{errortext}{red}. GeoPAVE preserves uncertainty, verifies candidate consistency, and avoids accepting over-specific coordinates that conflict with the visual scene.}
    \label{fig:qualitative_comparison}
\end{figure*}

Figure \ref{fig:qualitative_comparison} shows that our proposed model (GeoPAVE) produces reasoning trajectories with improved coherence and interpretability. In particular, the model keeps ambiguous visual cues as explicit hypotheses, records uncertainty when decisive text is absent, and uses external evidence to prevent visually plausible but over-specific candidate coordinates from being accepted.

We present a comparative example in Figure \ref{fig:qualitative_comparison} to illustrate the difference in reasoning capabilities. Given a suburban street view, \textbf{GPT-5.5} suffers from a severe cue-level error; it incorrectly identifies the scene as having right-hand traffic and misinterprets the road infrastructure as European, ultimately predicting Luxembourg with a distance error of over 8,700 km. \textbf{Gemini-3.1-Pro} demonstrates stronger perception by correctly identifying left-hand traffic, the distinctive interlocking zig-zag bricks, and the green palisade fencing, narrowing the location down to South Africa. However, it relies entirely on its internal parametric memory, leading to an incorrect city-level prediction (Pretoria). 

In contrast, \textbf{GeoPAVE} exhibits a structured reasoning trajectory. It first extracts fine-grained visual evidence, including the yellow road edge lines and interlocking brick median paving. More importantly, instead of making an immediate point prediction, our model formulates multiple hypotheses (Johannesburg vs. Pretoria) and leverages external tool-use (e.g., web queries for ``yellow road edge lines left hand drive countries'') to evaluate candidate consistency. When overly specific retrieved candidates conflict with the hilly sports-field scene, the evidence gate rejects them and falls back to the broader Johannesburg hypothesis.

\subsection{Failure Analysis}
\label{sec:failure_analysis}

Although GeoPAVE achieves the best overall performance on PAVED, its localization remains imperfect. Acc@200km is 69.35\%, compared with 85.86\% at 750 km (Table~\ref{tab:main_results}), indicating that fine-grained regional and city-level localization is still challenging. The two cases in Figure~\ref{fig:failure_cases} further illustrate how an early error can propagate through the reasoning process when the correct candidate is missing or external verification is unavailable.

\begin{itemize}
    \item \textbf{False textual grounding.} The ground-truth location is Melbourne, Australia ($-37.815567$, $144.964966$), whereas GeoPAVE predicts KICKZ Stachus in Munich, Germany ($48.138758$, $11.567166$), resulting in an error of 16,119.78 km. The visible ``KICKZ'' text is assigned excessive importance and treated as a location-specific anchor. Once Munich enters the candidate set, the subsequent support and verification stages search for evidence consistent with this hypothesis rather than testing whether the textual cue is geographically unique. The intended web queries also fail, leaving the model without independent evidence to reject the initial hypothesis.

    \item \textbf{Missing candidate coverage.} The ground-truth location is Moscow, Russia ($55.739385$, $37.584894$), but GeoPAVE predicts Sydney, Australia ($-33.862250$, $151.207684$), producing an error of 14,497.47 km. The night-time street scene contains little location-discriminative information, with a service-provider watermark serving as the only textual cue. Consequently, the initial hypothesis-generation stage produces only coarse city-level candidates and does not include Moscow. Misleading nearby-POI evidence then shifts the prediction toward Sydney, while failed web searches prevent the model from introducing or verifying alternatives outside the initial candidate set.
\end{itemize}

\end{document}